%% file: transfer.tex
\documentclass{article}

\usepackage[final]{nips_2017}

\usepackage[utf8]{inputenc} 
\usepackage[T1]{fontenc}    
\usepackage{hyperref}       
\usepackage{url}            
\usepackage{booktabs}       
\usepackage{amsfonts}       
\usepackage{nicefrac}       
\usepackage{microtype}      

\usepackage{times}
\usepackage{graphicx} 

\usepackage{algorithm}
\usepackage{algorithmic}

\usepackage{amsmath}               
{
	\newtheorem{assumption}{Assumption}
	
}

\usepackage{mdwlist}
\usepackage{xspace}
\usepackage{paralist}
\usepackage{color}

\usepackage{booktabs}
\usepackage{comment}

\usepackage{amssymb}
\usepackage{caption}
\usepackage{subcaption}

\usepackage{float}
\newfloat{algorithm}{t}{lop}

\usepackage{setspace}

\newcommand{\barnabs}{Barnab\'{a}s }
\newcommand{\poczos}{P\'{o}czos}

\newcommand{\argmin}{\mathrm{argmin}}

\usepackage{appendix}

\def\holder{H\"{o}lder }

\newtheorem{theorem}{Theorem}
\newtheorem{lemma}{Lemma}

\newcommand{\mat}[1]{\mathbf{#1}}

\newcommand{\norm}[1]{\left|\left|#1\right|\right|}
\newcommand{\abs}[1]{\left|#1\right|}

\newcommand*{\qed}{\hfill\ensuremath{\square}}%

 \usepackage[noindentafter]{titlesec}
 \titlespacing\section{0pt}{4pt plus 0pt minus 2pt}{2pt plus 0pt minus 1pt}

\title{Hypothesis Transfer Learning via \\Transformation Functions}

\author{
Simon S. Du\\
Carnegie Mellon University\\
\texttt{ssdu@cs.cmu.edu}\\
\And
Jayanth Koushik\\
   Carnegie Mellon University\\
\texttt{jayanthkoushik@cmu.edu}\\
\AND
Aarti Singh\\
Carnegie Mellon University\\
\texttt{aartisingh@cmu.edu}\\     
\And
\barnabs \poczos\\
Carnegie Mellon University\\
\texttt{bapoczos@cs.cmu.edu}\\ 
}

\begin{document}

\maketitle

\begin{abstract}
\label{sec:abs}
\input{abstract.tex}
\end{abstract}

\section{Introduction}

\label{sec:intro}
\input{intro.tex}

\section{Preliminaries}
\label{sec:pre}
\input{pre.tex}

\section{Transformation Functions}
\label{sec:hypo}
\input{hypo.tex}

\section{Excess Risk Analyses}
\label{sec:main}
\input{main.tex}

\section{Finding the Best Transformation Function}
\label{sec:safe}
\input{safe.tex}

\section{Experiments}
\label{sec:exp}
\input{exp.tex}

\section{Conclusion and Future Works}
\label{sec:con}
\input{conclusion.tex}

\section{Acknowledgements}
\label{sec:ack}
\input{ack.tex}

\bibliography{simonduref}
\bibliographystyle{plainnat}
\newpage
\appendix
\input{appendix.tex}

\end{document}

%% file: abstract.tex
We consider the Hypothesis Transfer Learning (HTL) problem where one incorporates a hypothesis trained on the source domain into the learning procedure of the target domain.
Existing theoretical analysis either only studies specific algorithms or only presents upper bounds on the generalization error but not on the excess risk.
In this paper, we propose a unified algorithm-dependent framework for HTL through a novel notion of \emph{transformation function}, which characterizes the relation between the source and the target domains.
We conduct a general risk analysis of this framework and 
in particular, we show for the first time, if two domains are related, HTL enjoys faster convergence rates of excess risks for Kernel Smoothing and Kernel Ridge Regression than those of the classical non-transfer learning settings.
Experiments on real world data demonstrate the effectiveness of our framework.


%% file: intro.tex
In a classical transfer learning setting, we have a large amount of data from a source domain and a relatively small amount of data from a target domain.
These two domains are related but not necessarily identical, and the usual assumption is that the hypothesis learned from the source domain is useful in the learning task of the target domain.

\begin{figure*}[t!]
    \centering
    \begin{subfigure}[t]{0.3\textwidth}
            \centering
            \includegraphics[width=0.8\textwidth]{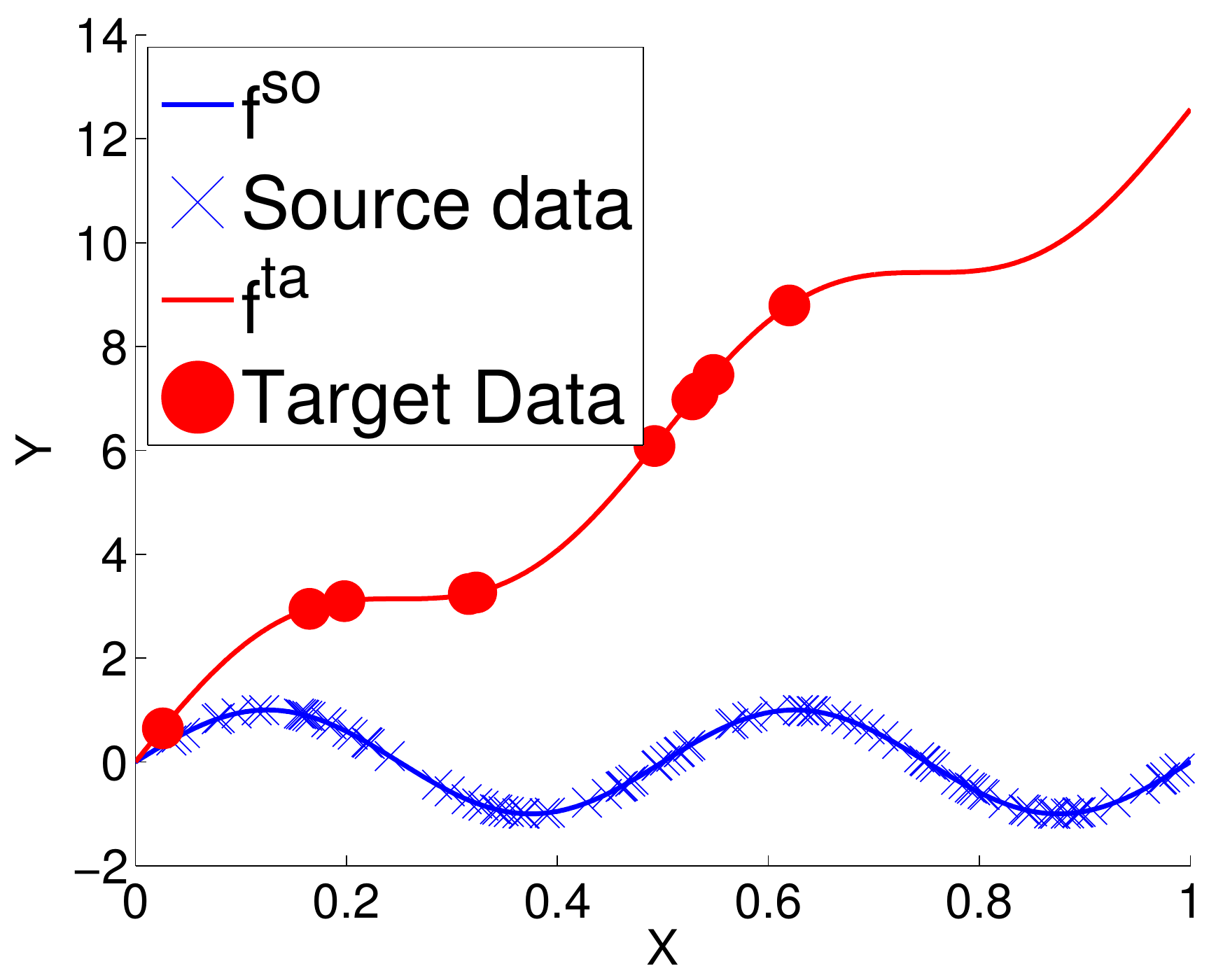}
            \caption{A toy example of transfer learning.
            We have many more samples from the source domain than the target domain.
            }
    \label{fig:model_shift_toy_example}
    \end{subfigure}
    \quad
    \begin{subfigure}[t]{0.3\textwidth}
        \centering
        \includegraphics[width=0.8\linewidth]{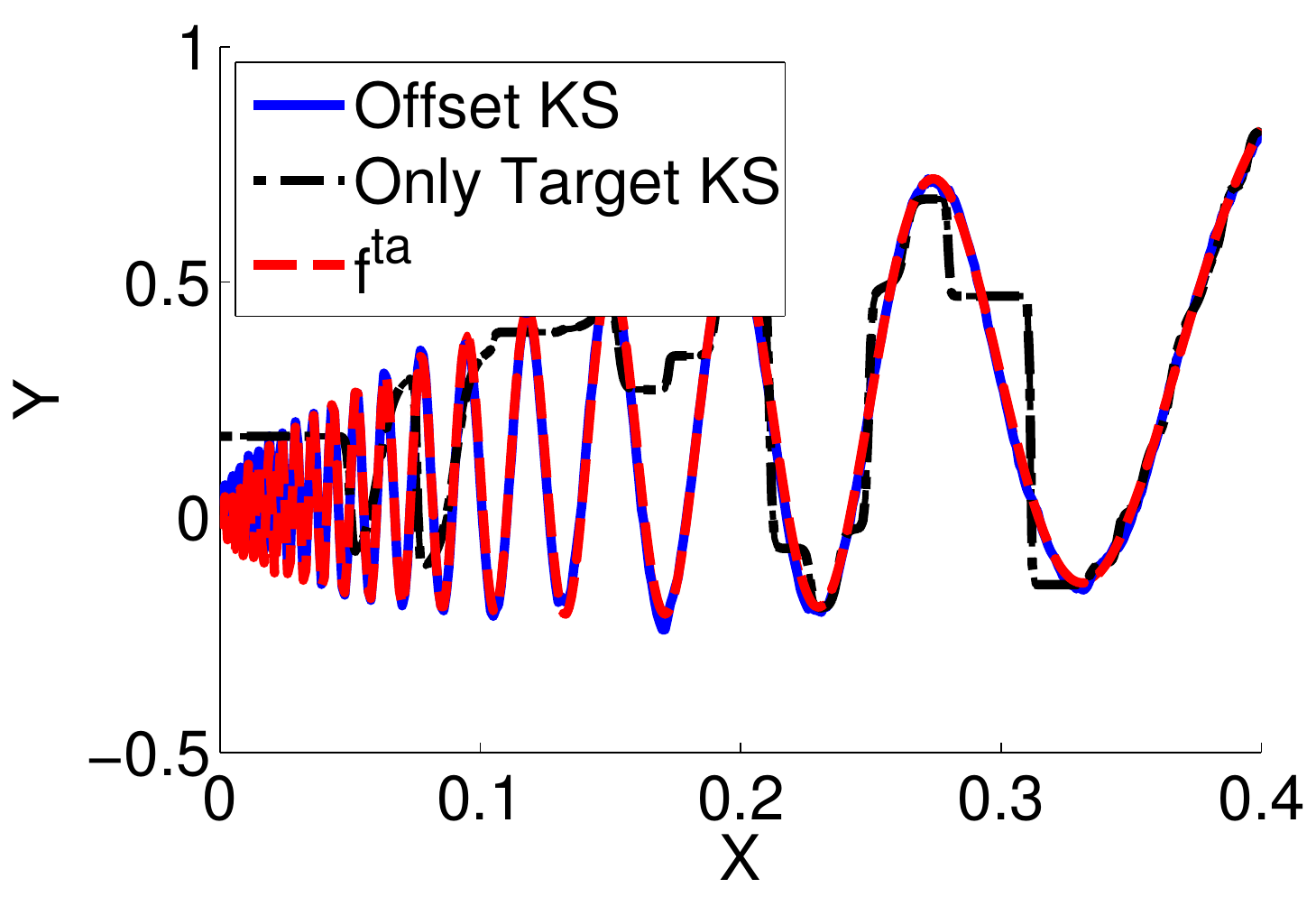}
        \caption{Transfer learning with Offset Transformation. }
        \label{fig:doppler_plus}
    \end{subfigure}
\quad
    \begin{subfigure}[t]{0.3\textwidth}
        \centering
        \includegraphics[width=0.8\linewidth]{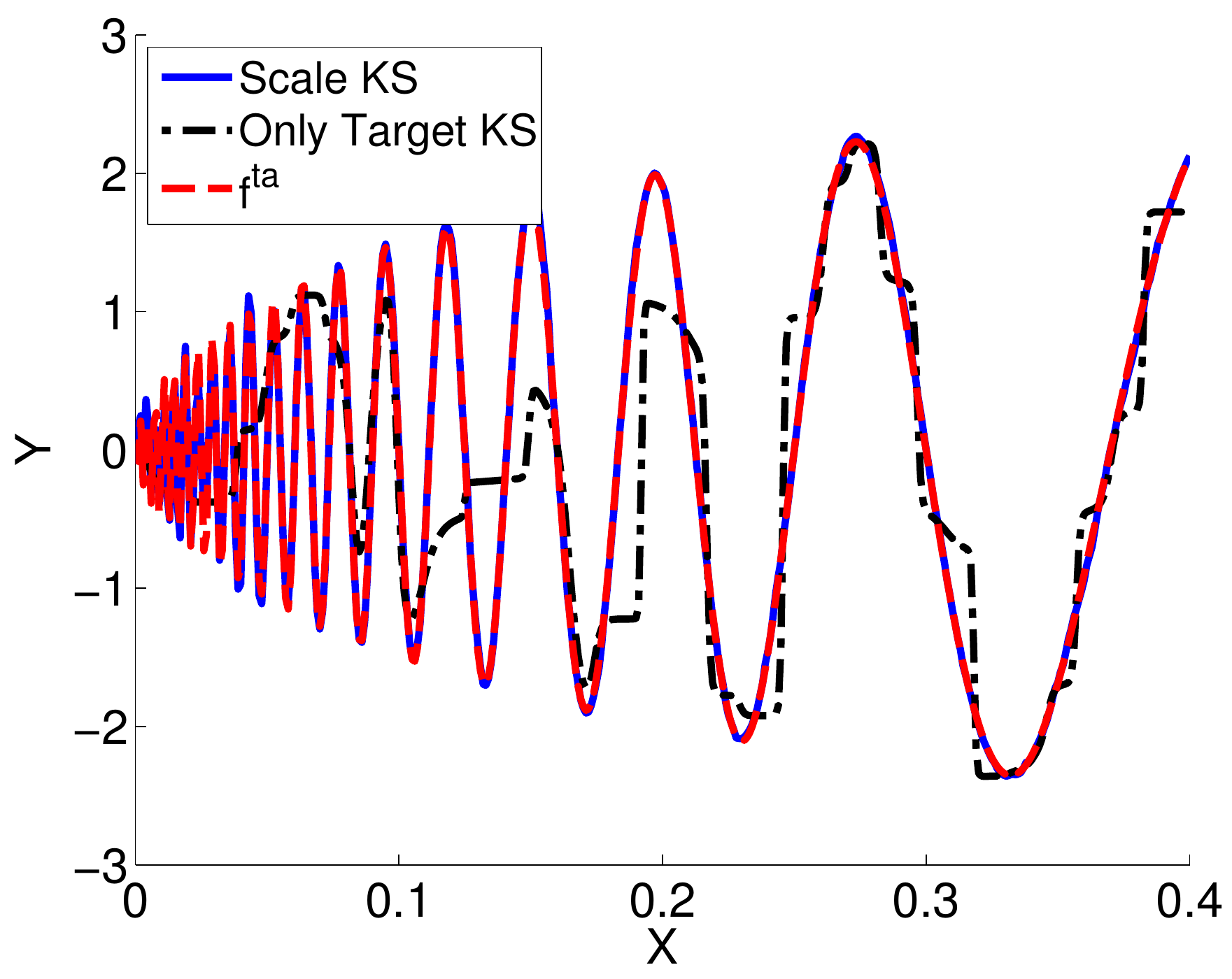}
        \caption{Transfer learning with Scale Transformation.}
        \label{fig:doppler_mult}
    \end{subfigure}
    \caption{Experimental results on synthetic data.}
    \label{fig:synthetic}
\end{figure*}

In this paper, we focus on the regression problem where the functions we want to estimate of the source and the target domains are different but related.
Figure~\ref{fig:model_shift_toy_example} shows a 1D toy example of this setting, where the source function is $f^{so}(x) = \sin(4\pi x)$ and the target function is $f^{ta}(x) = \sin(4\pi x) + 4\pi x$.
Many real world problems can be formulated as transfer learning problems.
For example, in the task of predicting the reaction time of an individual from his/her fMRI images, we have about $30$ subjects but each subject has only about $100$ data points.
To learn the mapping from neural images to the reaction time of a specific subject, we can treat all but this subject as the source domain, and this subject as the target domain.
In Section~\ref{sec:exp}, we show how our proposed method helps us learn this mapping more accurately.

This paradigm, hypothesis transfer learning (HTL) has been explored empirically with success in many applications~\citep{fei2006one,yang2007cross,orabona2009model,tommasi2010safety,kuzborskij2013n,wang2014flexible}.
\citet{kuzborskij2013stability,kuzborskij2016fast} pioneered the theoretical analysis of HTL for linear regression and recently~\citet{wang2015generalization} analyzed Kernel Ridge  Regression.
However, most existing works only provide generalization bounds, i.e. the difference between the true risk and the training error or the leave-one-out error.
These analyses are not complete because minimizing the generalization error does not necessarily reduce the true risk.
Further, these works often rely on a particular form of transformation from the source domain to the target domain.
For example, \citet{wang2015generalization} studied the offset transformation that instead of estimating the target domain function directly, they learn the residual between the target domain function and the source domain function.
It is natural to ask what if we use other transfer functions and how it affects the risk on the target domain. 

In this paper, we propose a general framework of HTL.
Instead of analyzing a specific form of transfer, we treat it as an input of our learning algorithm.
We call this input \emph{transformation function}  since intuitively, it captures the relevance between these two domains.\footnote{We formally define the transformation functions in Section~\ref{sec:hypo}.}
This framework unifies many previous works~\cite{wang2014flexible,kuzborskij2013stability,wang2016nonparametric} and naturally induces a class of new learning procedures.
Theoretically, we develop excess risk analysis for this framework.
The performance depends on the stability~\citep{bousquet2002stability}  of the algorithm used as a subroutine that if the algorithm is stable then the estimation error in the source domain will not affect the estimation in the target domain much.
To our knowledge, this connection was first established by~\citet{kuzborskij2013n} in the linear regression setting but here we generalize it to a broader context.
In particular, we provide explicit risk bounds for two widely used nonlinear estimators, Kernel Smoothing (KS) estimators and  Kernel Ridge Regression (KRR) as subroutines.
To the best of our knowledge, these are the first results showing when two domains are related, transfer learning techniques have faster statistical convergence rate of excess risk than that of non-transfer learning of kernel based methods.
Further, we accompany this framework with a theoretical analysis showing a small amount of data for cross-validation enables us (1) avoid using HTL when it is not useful and (2) choose the best transformation function as input from a large pool.
%

The rest of the paper is organized as follows. 
In Section~\ref{sec:pre} we introduce HTL and provide necessary backgrounds for KS and KRR.
We formalize our transformation function based framework in Section~\ref{sec:hypo}.
Our main theoretical results are in Section~\ref{sec:main} and specifically in Section~\ref{sec:ks} and Section~\ref{sec:krr} we provide explicit risk bounds for KS and KRR, respectively.
In Section~\ref{sec:safe} we analyze cross-validation in HTL setting and in Section~\ref{sec:exp} we conduct experiments on real world data data.
We conclude with a brief discussion of avenues for future work.


%% file: pre.tex
\subsection{Problem Setup}\label{sec:model_shift}
In this paper, we assume both $X \in \mathbb{R}^d$ and $Y \in \mathbb{R}$ lie in compact subsets: $\norm{X}_2 \le \triangle_X$, $\abs{Y} \le \triangle_Y$ for some $\triangle_X, \triangle_Y  \in \mathbb{R}_{+}$.
Throughout the paper, we use $\mathcal{T} = \left\{\left(X_i,Y_i\right)\right\}_{i=1}^n$ to denote a set of samples.
Let $\left(X^{so},Y^{so}\right)$ be the sample  from the source domain, and $\left(X^{ta},Y^{ta}\right)$ the sample from the target domain.
In our setting, there are $n_{so}$ samples drawn i.i.d from the source distribution: $\mathcal{T}^{so} = \left\{\left(X_i^{so},Y_i^{so}\right)\right\}_{i=1}^{n_{so}}$, and $n_{ta}$ samples drawn i.i.d from the target distribution: $\mathcal{T}^{ta} = \left\{\left(X_i^{ta},Y_i^{ta}\right)\right\}_{i=1}^{n_{ta}}$.
In addition, we also use $n_{val}$ samples drawn i.i.d from the target domain for cross-validation.
We model the joint relation between $X$ and $Y$ by:
$
Y^{so} = f^{so}\left(X^{so}\right) + \epsilon^{so}$ and
$Y^{ta} = f^{ta}\left(X^{ta}\right) + \epsilon^{ta}$ where $f^{so}$ and $f^{ta}$ are regression functions and we assume the noise $\mathbf{E}\left[\epsilon^{so}\right]=\mathbf{E}\left[\epsilon^{ta}\right] = 0$, i.i.d, and bounded.
We use $\mathcal{A}: \mathcal{T} \rightarrow \hat{f}$ to denote an algorithm that takes a set of samples and produce an estimator.
Given an estimator $\hat{f}$, we define the integrated $L_2$ risk as $R(\hat{f}) = \mathbf{E}\left[ \left(\hat{f}(X) - Y\right)^2 \right]$ where the expectation is taken over the distribution of $\left(X,Y\right)$.
%
%
Similarly, the empirical $L_2$ risk on a set of sample $\mathcal{T}$ is defined as  $
\hat{R}(\hat{f}) = \frac{1}{n}\sum_{i=1}^{n}\left(Y_i - \hat{f}\left(X_i\right)\right)^2.$
In HTL setting, we use $\hat{f}^{so}$ an estimator from the source domain to facilitate the learning procedure for $f^{ta}$.
\subsection{Kernel Smoothing}\label{sec:kernel_smoothing}
We say a function $f$ is in the
$\left(\lambda, \alpha\right)$ \holder class~\citep{wasserman2006all}, if for any $x, x' \in \mathbb{R}^d$, $f$ satisfies $
|f(x)-f(x')| \le \lambda \norm{x-x'}_2^{\alpha},
$ for some $\alpha \in \left(0,1\right)$.
The kernel smoothing method uses a positive kernel $K$ on $[0,1]$, highest at $0$, decreasing on $[0,1]$, $0$ outside $[0,1]$, and $\int_{\mathbb{R}^d}u^2K(u) < \infty$.
Using $\mathcal{T} = \left\{\left(X_i,Y_i\right)\right\}_{i=1}^n$, the kernel smoothing estimator is defined as follows:
$
\hat{f}(x) = \sum_{i=1}^{n}w_i(x)Y_i,
$ where $w_i(x) = \frac{K(\norm{x-X_i}/h)}{\sum_{j=1}^{n}K(\norm{x-X_j}/h)} \in [0,1].$
\subsection{Kernel Ridge Regression}
\label{sec:kernel_ridge_regression}
Another popular non-linear estimator is the kernel ridge regression (KRR) which uses the theory of reproducing kernel Hilbert space (RKHS) for regression~\citep{vovk2013kernel}.
Any symmetric positive semidefinite kernel function $K: \mathcal{R}^d \times \mathcal{R}^d \rightarrow \mathcal{R}$ defines a RKHS $\mathcal{H}$.
For each $x \in \mathcal{R}^d$, the function $z \rightarrow K(z,x)$ is contained in the Hilbert space $\mathcal{H}$; moreover, the Hilbert space is endowed with an inner product $\langle\cdot,\cdot\rangle _\mathcal{H}$ such that $K(\cdot, x)$ acts as the kernel of the evaluation functional, meaning
$
\langle f,K(x,\cdot)\rangle_\mathcal{H} = f(x) \text{ for } f \in \mathcal{H}.
$
In this paper we assume $K$ is bounded: $\sup_{x \in \mathbb{R}^{d}}K\left(x,x\right) = k < \infty$.
Given the inner product, the $\mathcal{H}$ norm of a function $g \in \mathcal{H}$ is defined as $\norm{g}_\mathcal{H} \triangleq \sqrt{\langle g,g\rangle}_\mathcal{H}$ and similarly the $L_2$ norm, $\norm{g}_2 \triangleq \left(\int_{\mathbb{R}^d} g(x)^2  dP_X\right)^{1/2}$ for a given $P_X$.
Also, the kernel induces an integral operator $T_K: L_2\left(P_X\right) \rightarrow L_2\left(P_X\right)$: $
T_K\left[f\right]\left(x\right) = \int_{\mathbb{R}^d} K\left(x',x\right)f\left(x'\right)dP_x\left(x'\right)
$ with countably many non-zero eigenvalues: $\left\{\mu_i\right\}_{i\ge1}$.
For a given function $f$, the approximation error is defined as: $
A_{f}\left(\lambda\right) \triangleq \inf_{h \in \mathcal{H}} \left(\norm{h-f}_{L_2\left(P_X\right)}^2+\lambda\norm{h}_\mathcal{H}^2 \right)
$ for $\lambda \ge 0$.
Finally the estimated function evaluated at point $x$ can be written as $
\hat{f}\left(x\right) = K(\mat{X},x)\left(K(\mat{X},\mat{X}) + n\lambda I\right)^{-1}\mat{Y}
$ where $\mat{X} \in \mathbb{R}^{n \times d}$ are the inputs of training samples and $\mat{Y} \in \mathbb{R}^{n \times 1}$ are the training labels~\cite{vovk2013kernel}.

\subsection{Related work}
Before we present our framework, it is helpful to give a brief overview of existing literature on theoretical analysis of transfer learning.
Many previous works focused on the settings when only \emph{unlabeled} data from the target domain are available~\citep{huang2006correcting,sugiyama2008direct,yu2012analysis}.
In particular, a line of research has been established based on distribution discrepancy, a loss induced metric for the source and target distributions~\citep{mansour2009domain,ben2007analysis,blitzer2008learning,cortes2011domain,mohri2012new}.
For example, recently~\citet{cortes2014domain} gave generalization bounds for kernel based methods under convex loss in terms of discrepancy.

In many real world applications such as yield prediction from pictures~\citep{nuske2014modeling}, or prediction of response time from fMRI~\citep{verstynen2014organization}, some labeled data from the target domain is also available.
\citet{cortes2015adaptation} used these data to improve their discrepancy minimization algorithm.
\citet{zhang2013domain} focused on modeling target shift ($P(Y)$ changes), conditional shift ($P(X|Y)$ changes), and a combination of both.
Recently,~\citet{wang2014flexible} proposed a kernel mean embedding method to match the conditional probability in the kernel space and later derived generalization bound for this problem~\cite{wang2015generalization}.
~\citet{kuzborskij2013stability,kuzborskij2016fast,kuzborskij2016scalable} gave excess risk bounds for target domain estimator in the form of
a linear combination of estimators from multiple source domains and an additional linear function.
\citet{ben2013domain} showed a similar bound of the same setting with different quantities capturing the relatedness.
\citet{wang2016nonparametric} showed that if the features of source and target domain are $[0,1]^d$, using orthonormal basis function estimator, transfer learning achieves better excess risk guarantee if $f^{ta}-f^{so}$ can be approximated by the basis functions easier than $f^{ta}$. 
Their work can be viewed as a special case of our framework using the transformation function $G(a,b) = a+b$.

%% file: hypo.tex
In this section, we first define our class of models and give a meta-algorithm to learn the target regression function.
Our models are based on the idea that transfer learning is helpful when one transforms the target domain regression problem into a simpler regression problem using source domain knowledge.
Consider the following example.\\
\textbf{Example: Offset Transfer.} 
Let 
$
f^{so}(x) = \sqrt{x\left(1-x\right)}\sin\left(\frac{2.1\pi}{x+0.05}\right)$ and
$f^{ta}(x) = f^{so}(x) + x. \label{eqn:doppler}
$
$f^{so}$ is the so called Doppler function. 
It requires a large number of samples to estimate well because of its lack of smoothness~\cite{wasserman2006all}.
For the same reason, $f^{ta}$ is also difficult to estimate directly.
However, if we have enough data from the source domain, we can have a fairly good estimate of $f^{so}$.
Further, notice that the offset function $w(x) = f^{ta}(x)-f^{so}(x) = x$, is just a linear function.
Thus, instead of directly using $\mathcal{T}^{ta}$ to estimate $f^{ta}$, we can use the target domain samples to find an estimate of $w(x)$, denoted by $\hat{w}(x)$, and our estimator for the target domain is just: $\hat{f}^{ta}(x) = \hat{f}^{so}(x) + \hat{w}(x)$.
Figure~\ref{fig:doppler_plus} shows this technique gives improved fitting for $f^{ta}$.

The previous example exploits the fact that function $w(x)=f^{ta}(x)-f^{so}(x)$ is a simpler function than $f^{ta}$.
Now we generalize this idea further.
Formally, we define the \emph{transformation function} as 
$
G(a,b): \mathbb{R}^2 \rightarrow \mathbb{R}
$ where we assume that given $a \in \mathbb{R}$, $G(a,\cdot)$ is invertible.
Here $a$ will be the regression function of the source domain evaluated at some point and the output of $G$ will be the regression function of the target domain evaluated at the same point.
Let $G_a^{-1} (\cdot)$ denote the inverse of $G(a,\cdot)$ such that $
G\left(a,G_a^{-1}\left(c\right)\right) = c.
$
For example if $G(a,b) = a+b$ and $G^{-1}_{a}\left(c\right) = c-a$.
For a given $G$ and a pair $\left(f^{so},f^{ta}\right)$, they together induce a function $
w_G(x) = G_{f^{so}\left(x\right)}^{-1} (f^{ta}(x)).
$
In the offset transfer example, $w_G\left(x\right) = x$.
By this definition, for any $x$, we have $
G\left(
f^{so}\left(x\right), w_G\left(x\right)\right) = f^{ta}\left(x\right)
.
$
We call $w_G$ the \emph{auxiliary function} of the transformation function $G$.  
In the HTL setting, $G$ is a user-defined transformation that represents users' prior knowledge on the relation between the source and target domains.
Now we list some other examples:\\
\textbf{Example: Scale-Transfer.} Consider $G(a,b) = ab$. 
This transformation function is useful when $f^{so}$ and $f^{ta}$ satisfy a smooth scale transfer.
For example, if $f^{ta} = cf^{so}$, for some constant $c$, then $w_G(x)=c$ because $f^{ta}\left(x\right) = G\left(f^{so}\left(x\right),w_G\left(x\right)\right) = f^{so}\left(x\right)w_G\left(x\right)=f^{so}\left(x\right)c$.
See Figure~\ref{fig:doppler_mult}.\\
\textbf{Example: Non-Transfer.} Consider $G(a,b) = b$. 
Notice that $f^{ta}(x) = w_G(x)$ and so $f^{so}$ is irrelevant.
Thus this model is equivalent to traditional regression on the target domain since data from the source domain does not help.


\subsection{A Meta Algorithm}\label{sec:algo}
Given the transformation $G$ and data, we provide a general procedure to estimate $f^{ta}$.
The spirit of the algorithm is turning learning a complex function $f^{ta}$ into an easier function $w_G$.
First we use an algorithm $\mathcal{A}_{so}$ that takes $\mathcal{T}^{so}$ to obtain $\hat{f}^{so}$.
Since we have sufficient data from the source domain, $\hat{f}^{so}$ should be close to the true regression function $f^{so}$.
Second, we construct a new data set using the $n_{ta}$ data points from the target domain: $\mathcal{T}^{w_G} = \left\{\left(X_i^{ta},H_G\left(\hat{f}^{so}\left(X_i^{ta}\right), Y_i^{ta}\right)\right)\right\}_{i=1}^{n_{ta}}$ where  $H_G: \mathbb{R}^2 \rightarrow \mathbb{R}$ and satisfies 
\[ \mathbf{E}\left[H_G\left(f^{so}\left(X_i^{ta}\right),Y_i^{ta}\right)\right] =  
G^{-1}_{f^{so}\left(X_i^{ta}\right)}\left(f^{ta}\left(X_i^{ta}\right)\right)
= w_G\left(X_i^{ta}\right)
\]
where and the expectation is taken over $\epsilon_{ta}$.
Thus, we can use these newly constructed data to learn $w_G$ with algorithm $\mathcal{A}_{W_G}$: $\hat{w}_G = \mathcal{A}_{W_G}\left(\mathcal{T}^{W_G}\right)$.
Finally, we plug trained $\hat{f}^{so}$ and $\hat{w}_G$ into transformation $G$ to obtain an estimation for $f^{ta}$:
$\hat{f}^{ta}(X) = G(\hat{f}^{so}\left(X\right),\hat{w}_G(X)).$ 
Pseudocode is shown in Algorithm~\ref{algo:generalForm}.

\paragraph{Unbiased Estimator $H_G\left(f^{so}\left(X^{ta}\right),Y^{ta}\right)$:}
In Algorithm~\ref{algo:generalForm}, we require an unbiased estimator for $w_G\left(X^{ta}\right)$.
Note that if $G\left(a,b\right)$ is linear $b$ or $\epsilon^{ta} = 0$, we can simply set $H_G\left(f^{so}\left(X\right),Y\right) = G^{-1}_{f^{so}\left(X\right)}\left(Y\right)$.
For other scenarios, $G^{-1}_{f^{so}\left(X_i^{ta}\right)}\left(Y_i^{ta}\right)$ is biased: $\mathbf{E}\left[G^{-1}_{f^{so}\left(X_i^{ta}\right)}\left(Y_i^{ta}\right)\right] \neq G_{f^{so}\left(x\right)}^{-1}\left(f^{ta}\left(x\right)\right)$ and we need to design estimator using the structure of $G$.\\
\textbf{Remark 1:}
Many transformation functions are equivalent to a transformation function $G'\left(a,b\right)$ where $G'\left(a,b\right)$ is linear in $b$.
For example, for $G\left(a,b\right) = ab^2$, i.e., $f^{ta}\left(x\right) = f^{so}\left(x\right)w_G^2\left(x\right)$, consider $G'\left(a,b\right)=ab$ where $b$ in $G'$ stands for $b^2$ in $G$, i.e., $f^{ta}\left(x\right) = f^{so}\left(x\right) w_G'\left(x\right)$.
Therefore $w_G' = w_G^2$ and we only need to estimate $w_G'$ well instead of estimating $w_G$.
More generally, if $G\left(a,b\right)$ can be factorized as $G\left(a,b\right) = g_1\left(a\right)g_2\left(b\right)$, i.e., $f^{ta}\left(x\right) = g_1\left(f^{so}\left(x\right)\right)g_2\left(w_G\left(x\right)\right)$, we only need to estimate $g_2\left(w_G\left(x\right)\right)$ and the convergence rate depends on the structure of $g_2\left(w_G\left(x\right)\right)$.\\
\textbf{Remark 2:}
When $G$ is not linear in $b$ and $\epsilon^{ta} \neq 0$, observe that in Algorithm~\ref{algo:generalForm}, we treat $Y_i^{ta}$s as noisy covariates to estimate $w_G\left(X_i\right)$s.
This problem is called error-in-variable or measurement error and has been widely studied in statistics literature.
For details, we refer the reader to the seminal work by~\citet{carroll2006measurement}.
There is no universal estimator for the measurement error problem.
In Section~\ref{sec:measure_err}, we provide a common technique, regression calibration to deal with measurement error problem.

\begin{algorithm}[tb]
	\renewcommand{\algorithmicrequire}{\textbf{Inputs:}}
	\renewcommand{\algorithmicensure}{\textbf{Outputs:}}
	\caption{Transformation Function based Transfer Learning}
	\label{algo:generalForm}
	\begin{algorithmic}[1]
		\REQUIRE Source domain data: $\mathcal{T}^{so} = \{\left(X^{so}_i,Y^{so}_i\right)\}_{i=1}^{n_{so}}$, target domain data: $\mathcal{T}^{ta} = \{\left(X^{ta}_i,Y^{ta}_i\right)\}_{i=1}^{n_{ta}}$, transformation function: $G$, algorithm to train $f^{so}$: $\mathcal{A}_{so}$, algorithm to train $w_G$: $\mathcal{A}_{w_G}$ and $H_G$ an unbiased estimator for estimating $w_G$. 
		\ENSURE Regression function for the target domain: $\hat{f}^{ta}$.
		\STATE Train the source domain regression function $\hat{f}^{so} = \mathcal{A}_{so}\left(\mathcal{T}^{so}\right)$.
		\STATE Construct new data using $\hat{f}^{so}$ and $\mathcal{T}^{ta}$: $\mathcal{T}^{w_G} = \left\{\left(X_i^{ta},W_i\right)\right\}_{i=1}^{n_{ta}}$, where $W_i = H_G\left(\hat{f}^{so}\left(X_i^{ta}\right),Y_i^{ta}\right)$.
		\STATE Train the auxiliary function: $\hat{w}_G = \mathcal{A}_{W_G}\left(\mathcal{T}^{w_G}\right)$.
		\STATE Output the estimated regression for the target domain: $\hat{f}^{ta}(X) = G\left(\hat{f}^{so}(X), \hat{w}_G(X)\right)$.
	\end{algorithmic}
\end{algorithm}

%% file: main.tex
In this section, we present theoretical analyses for the proposed class of models and estimators.
First, we need to impose some conditions on $G$.
The first assures that if the estimations of $f^{so}$ and $w_G$ are close to the source regression and auxiliary function, then our estimator for $f^{ta}$ is close to the true target regression function. 
The second assures that we are estimating a regular function.
\begin{assumption}\label{assum:G_lipschitz}
 $G\left(a,b\right)$ is $L$-Lipschitz: $|G(a,b) - G(a',b')| \le L \norm{(a,b)-(a',b')}_2$ and is invertible with respect to $b$ given $a$, i.e. if $G\left(x,y\right) = z$ then $G_x^{-1}\left(z\right) = y$.
\end{assumption}
\begin{assumption}\label{assum:w_bounded}
Given $G$, the induced auxiliary function $w_G$ is bounded: for $x: \norm{x}_2\le\triangle_X$, $w_G\left(x\right) \le B$ for some $B > 0$.
\end{assumption}
Offset Transfer and Non-Transfer satisfy these conditions with $L=1$ and $B = \triangle_Y$. 
Scale Transfer satisfies these assumptions when $f^{so}$ is lower bounded from away $0$.
Lastly, we assume our unbiased estimator is also regular.
\begin{assumption}\label{assum:G_inverse_lipschitz}
For $x:\norm{x}_2\le \triangle_X$ and $y:\abs{y}\le \triangle_Y$, $H_G\left(x,y\right)\le B$ for some $B > 0$ and $H_G$ is Lipschitz continuous in the first argument:$
\abs{H_G\left(x,y\right)-H_G\left(x',y\right)} \le L\abs{x-x'}
$ for some $L > 0$.
\end{assumption}

We begin with a general result which only requires the stability of $\mathcal{A}_{W_G}$:
\begin{theorem}\label{thm:general_thm}
Suppose for any two sets of samples that have same features but different labels: $\mathcal{T} = \left\{\left(X_i^{ta},W_i\right)\right\}_{i=1}^{n_{ta}}$ and $\widetilde{\mathcal{T}} =\left\{\left(X_i^{ta},\widetilde{W}_i\right)\right\}_{i=1}^{n_{ta}}$, 
the algorithm $\mathcal{A}_{w_G}$ for training $w_G$ satisfies:
\begin{align}
\norm{\mathcal{A}_{w_G}\left(\mathcal{T}\right) - \mathcal{A}_{w_G}\left(\widetilde{\mathcal{T}}\right)}_{\infty} \le \sum_{i=1}^{n_{ta}} c_i\left(X_i^{ta}\right)\left|W_i - \widetilde{W}_i\right|, \label{eqn:stability}
\end{align} where $c_i$ only depends on $X_i^{ta}$.
Then for any $x$,
\begin{align*}
\left|\hat{f}^{ta}(x)-f^{ta}(x)\right|^2  =  & O\left(
\left|\hat{f}^{so}\left(x\right)-f^{so}\left(x\right)\right|^2 + \left|\widetilde{w}_G\left(x\right) - w_G\left(x\right)\right|^2
+ \right. \\ 
& \left.   \left(\sum_{i=1}^{n_{ta}}c_i\left(X_i^{ta}\right)\left|\hat{f}^{so}\left(X_i^{ta}\right)-f^{so}\left(X_i^{ta}\right)\right|\right)^2
\right)
\end{align*}where $\tilde{w}_G = \mathcal{A}_{w_G}\left(\left\{\left(X_i^{ta}, H_G\left(f^{so}\left(X_i^{ta}\right),Y_i^{ta}\right)\right)\right\}_{i=1}^{n_{ta}}\right)$, the estimated auxiliary function trained based on true source domain regression function.
\end{theorem}

Theorem~\ref{thm:general_thm} shows how the estimation error in the source domain function propagates to our estimation of the target domain function.
Notice that if we happen to know $f^{so}$, then the error is bounded by $O\left(\abs{\tilde{w}_G\left(x\right) - w_G\left(x\right)}^2\right)$, the estimation error of $w_G$.
However, since we are using estimated $f^{so}$ to construct training samples for $w_G$, the error might accumulate as $n_{ta}$ increases.
Though the third term in Theorem~\ref{thm:general_thm} might increase with $n_{ta}$, it also depends on the estimation error of $f^{so}$ which is relatively small because of the large amount of source domain data.

The stability condition~\eqref{eqn:stability} we used is related to the uniform stability introduced by Bousquet and Elisseeff~\cite{bousquet2002stability} where they consider how much will the output change if one of the training instance is removed or replaced by another whereas ours depends on two different training data sets.
The connection between transfer learning and stability has been discovered by~\citet{kuzborskij2013stability,liu2016algorithm} and ~\citet{zhang2015multi} in different settings, but they only showed bounds for generalization, not for excess risk.
\subsection{Kernel Smoothing}\label{sec:ks}
We first analyze kernel smoothing method.
\begin{theorem}\label{thm:kernel_smoothing_global}
Suppose the support of $X^{ta}$ is a subset of the support of $X^{so}$ and the probability density of $P_{X^{so}}$ and $P_{X^{ta}}$ are uniformly bounded away from below on their supports.
Further assume $f^{so}$ is $(\lambda_{so},\alpha_{so})$ \holder and $w_G$ is $(\lambda_{w_G},\alpha_{w_G})$ \holder. 
If we use kernel smoothing estimation for $f^{so}$ and $w_G$ with bandwidth $h_{so} \asymp n_{so}^{-1/(2\alpha_{so}+d)}$ and $h_{w_G} \asymp n_{ta}^{-1/(2\alpha_{w_G}+d)}$, with probability at least $1-\delta$ the risk satisfies:
\begin{align*}
\mathbf{E}\left[R \left(\hat{f}^{ta}\right)\right] - R \left(f^{ta}\right)  =
O\left(n_{so}^{\frac{-2\alpha_{so}}{2\alpha_{so}+d}} + n_{ta}^{\frac{-2\alpha_{w_G}}{2\alpha_{w_G}+d}}\right)\log\left(\frac{1}{\delta}\right)
\end{align*} where the expectation is taken over $\mathcal{T}^{so}$ and $\mathcal{T}^{ta}$.
\end{theorem}
Theorem~\ref{thm:kernel_smoothing_global} suggests that the risk depends on two sources, one from estimation of $f^{so}$ and one from estimation of $w_G$.
For the first term, since in  the typical transfer learning scenarios $n_{so} >> n_{ta}$, it is relatively small in the setting we focus on.
The second terms shows the power of transfer learning on transforming a possibly complex target regression function into a simpler auxiliary function.
It is well known that learning $f^{ta}$ only using target domain has risk of the order $\Omega\left(n_{ta}^{-2\alpha_{f^{ta}}/\left(2\alpha_{f^{ta}}+d\right)}\right)$.
Thus, if the auxiliary function is smoother than the target regression function, i.e. $\alpha_{w_G} > \alpha_{f^{ta}}$, we obtain better statistical rate.

\subsection{Kernel Ridge Regression}
\label{sec:krr}
Next, we give an upper bound for the excess risk using KRR:
\begin{theorem}\label{thm:RKHS_regression}
Suppose $P_{X^{so}} = P_{X^{ta}}$ and  the eigenvalues of the integral operator $T_K$ satisfy $
\mu_i \le a i^{-1/p}$ for $\qquad i \ge 1$
$a \ge 16\triangle_Y^4$, $p \in \left(0,1\right)$ and there exists a constant $C \ge 1$ such that for $f \in \mathcal{H}$, $\norm{f}_\infty \le C \norm{f}_\mathcal{H}^p \cdot \norm{f}_{L_2\left(P_X\right)}^{1-p}$.
Furthur assume that $A^{f^{so}}\left(\lambda\right) \le c\lambda^{\beta_{so}}$ and $A^{w_{G}}\left(\lambda\right) \le c\lambda^{\beta_{w_G}}$.
If we use KRR for estimating $f^{so}$ and $w_G$ with regularization parameters $\lambda_{so} \asymp n_{so}^{-1/\left(\beta_{so}+p\right)}$ and $\lambda_{w_G}  \asymp n_{ta}^{-1/(\beta_{w_G}+p)}$, then with probability at least $1-\delta$ the excess risk satisfies:
\begin{align*}
\mathbf{E}\left[R \left(\hat{f}^{ta}\right)\right] - R \left(f^{ta}\right)  =  O\left(\left(n_{ta}^{\frac{2}{\beta_{w_G}+p}}\log\left(n_{ta}\right)\cdot n_{so}^{\frac{-\beta_{so}}{\beta_{so}+p}} + n_{ta}^{\frac{-\beta_{w_G}}{\beta_{w_G}+p}}\right)\log\left(\frac{1}{\delta}\right)\right)
\end{align*}where the expectation is taken over $\mathcal{T}^{so}$ and $\mathcal{T}^{ta}$.
\end{theorem}

Similar to Theorem~\ref{thm:kernel_smoothing_global}, Theorem~\ref{thm:RKHS_regression} suggests that the estimation error comes from two sources.
For estimating the auxiliary function $w_G$, the statistical rate depends on properties of the kernel induced RKHS, and how far the auxiliary function is from this space.
For the ease of presentation, we assume $P_{X^{so}} = P_{X^{ta}}$, so the approximation errors $A^{f^{so}}$ and $A^{f^{ta}}$ are defined on the same domain.
The error of estimating $f^{so}$ is amplified by $O\left(\lambda_{w_G}^{-2}\log\left(n_{ta}\right)\right)$, which is worse than that of nonparametric kernel smoothing.
We believe this $\lambda_{w_G}^{-2}$ is nearly tight because Bousquet and Elisseeff have shown the uniform algorithmic stability parameter for KRR is $O\left(\lambda_{w_G}^{-2}\right)$~\cite{bousquet2002stability}.
Steinwart et al.~\cite{steinwart2009optimal} showed that for non-transfer learning, the optimal statistical rate for excess risk is $\Omega\left(n_{ta}^{\frac{-\beta_{ta}}{\beta_{ta}+p}}\right)$, so if $\beta_{wg} \ge \beta_{ta}$ and $n_{so}$ is sufficiently large then we achieve improved convergence rate through transfer learning.

\textbf{Remark:} Theorem~\ref{thm:kernel_smoothing_global} and~\ref{thm:RKHS_regression} are not directly comparable because our assumptions on the function spaces of these two theorems are different. 
In general, \holder space is only a Banach space but not a Hilbert space.
We refer readers to Theorem 1 in~\cite{zhou2008derivative} for details.

%% file: safe.tex
In the previous section we showed for a specific transformation function $G$, if auxiliary function is smoother than the target regression function then we have smaller excess risk.
In practice, we would like to try out a class of transformation functions $\mathcal{G}$
, which is possibly uncountable.
We can construct a subset of $\mathcal{\overline{G}} \subset \mathcal{G}$, which is finite and satisfies that each $G$ in $\mathcal{G}$ there is a $\overline{G}$ in $\mathcal{\overline{G}}$ that is close to $G$.
Here we give an example. 
Consider the transformation functions that have the form:
$
\mathcal{G} = \left\{G(a,b) = \alpha a + b \text{ where } \abs{\alpha}\le L_\alpha, \abs{a}\le L_a \right\}.
$
We can quantize this set of transformation functions by considering a subset of 
$\mathcal{G}$:
$
\overline{\mathcal{G}} = \left\{G(a,b) = k\epsilon a + b \right\} 
 \text{ where } \epsilon = \frac{L_\alpha}{2K}, k = -K,\cdots,0,\cdots, K \text{ and } \abs{a}\le L_a .
$
Here $\epsilon$ is the quantization unit.

The next theorem shows 
we only need to search the transformation function $\overline{G}$ in $\overline{\mathcal{G}}$ whose corresponding estimator $\hat{f}^{ta}_{\overline{G}}$ has the lowest empirical risk on the validation dataset.
\begin{theorem}\label{thm:best_in_class}
Let $\mathcal{G}$ be a class of transformation functions and $\overline{\mathcal{G}}$ be its $\norm{\cdot}_\infty$ norm $\epsilon$-cover.
Suppose $w_G$ satisfies the same assumption in Theorem~\ref{thm:general_thm} and for any two $G_1, G_2 \in \mathcal{G}$, 
$\norm{w_{G_1}-w_{G_2}}_\infty \le L\norm{G_1 - G_2}_\infty$ for some constant $L$.
Denote $
G^\star = \argmin_{G\in\mathcal{G}}R\left(\hat{f}^{ta}_{G}\right)$  and  $\overline{G}^\star = \argmin_{G \in \overline{\mathcal{G}}} \hat{R}\left(\hat{f}^{ta}_{G}\right)$.
If  we choose $\epsilon = O\left(\frac{R\left(\hat{f}^{ta}_{G^\star}\right)}{\sum_{i=1}^{n_{ta}}ci}\right)$ and $n_{val} = \Omega\left(\log\left(\abs{\overline{\mathcal{G}}}/\delta\right)\right)$, the with probability at least $1-\delta$, $\mathbf{E}\left[R\left(\hat{f}^{ta}_{\overline{G}^\star}\right)\right] - R\left(f^{ta}\right) = O\left(\mathbf{E}\left[R\left(\hat{f}^{ta}_{G^\star}\right) \right]- R\left(f^{ta}\right)\right)$ where the expectation is taken over $\mathcal{T}^{so}$ and $\mathcal{T}^{ta}$.
\end{theorem} 
\paragraph{Remark 1:} This theorem implies that if no-transfer function ($G\left(a,b\right)=b$) is in $\mathcal{G}$ then we will end up choosing a transformation function that has the same order of excess risk as using no-transfer learning algorithm, thus avoiding negative transfer.
\paragraph{Remark 2:} Note number of validation set is only logarithmically depending on the size of set of transformation functions.
Therefore, we only need to use a very small amount of data from the target domain to do cross-validation.

%% file: exp.tex
In this section we use robotics and neural
imaging data to demonstrate the effectiveness of the proposed framework.
We conduct experiments on real-world data sets with the following procedures.
\begin{itemize*}
\item Directly training on the target data $\mathcal{T}^{ta}$ (Only Target KS, Only Target KRR).
\item Only training on the source data $\mathcal{T}^{so}$ (Only Source KS, Only Source KRR).
\item Training on the combined source and target data (Combined KS, Combined KRR).
\item The CDM algorithm proposed by~\citet{wang2014flexible} with KRR (CDM).
\item The algorithm described in this paper with $G(a,b) = (a+\alpha)b$ where $\alpha$ is a hyper-parameter (Scale KS, Scale KRR).
\item The algorithm described in this paper with $G(a,b) = \alpha a+b$ where $\alpha$ is a hyper-parameter (Offset KS, Offset KRR).
$\angle$

\end{itemize*}

\begin{table*}
    \centering
    \resizebox{1\columnwidth}{!}{%
    \begin{tabular}{@{}lllllll@{}}
        \toprule
        & $n_{ta} = 10$ & $n_{ta} = 20$ & $n_{ta} = 40$ & $n_{ta} = 80$ &
            $n_{ta} = 160$ & $n_{ta} = 320$\\
        \midrule
        Only Target KS & $0.086 \pm 0.022$ & $0.076 \pm 0.010$ & $0.066
            \pm 0.008$ & $0.064 \pm 0.007$ & $0.065 \pm 0.006$ & $0.063 \pm 0.005$\\
        Only Target KRR & $0.080 \pm 0.017$ & $0.078 \pm 0.022$ & $0.063
            \pm 0.013$ & $0.050 \pm 0.007$ & $0.048 \pm 0.006$ &
            $\textbf{0.040}~\mathbf{\pm}~\textbf{0.005}$\\
        Only Source KRR & $0.098 \pm 0.017$ & $0.098 \pm 0.017$ & $0.098 \pm
            0.017$ & $0.098 \pm 0.017$ & $0.098 \pm 0.017$ & $0.098 \pm 0.017$\\
        Combined KS & $0.092 \pm 0.011$ & $0.084 \pm 0.008$ & $0.077
            \pm 0.009$ & $0.075 \pm 0.006$ & $0.074 \pm 0.006$ & $0.067 \pm 0.006$\\
        Combined KRR & $0.087 \pm 0.025$ & $0.077 \pm 0.015$ & $0.062 \pm
            0.009$ & $0.061 \pm 0.005$ & $0.047 \pm 0.003$ & $0.041 \pm 0.004$\\
        CDM & $0.105 \pm 0.023$ & $0.074 \pm 0.020$ & $0.064 \pm 0.008$
            & $0.060 \pm 0.007$ & $0.053 \pm 0.009$ & $0.056 \pm 0.004$\\
        Offset KS & $0.080 \pm 0.026$ & $0.066 \pm 0.023$ &
            $\textbf{0.052}~\mathbf{\pm}~\textbf{0.006}$ & $0.054 \pm 0.006$ &
            $0.050 \pm 0.003$ & $0.052 \pm 0.004$\\
        Offset KRR & $0.146 \pm 0.112$ & $0.066 \pm 0.017$ & $0.053 \pm
            0.007$ & $\textbf{0.048}~\mathbf{\pm}~\textbf{0.006}$ &
            $\textbf{0.043}~\mathbf{\pm}~\textbf{0.004}$ & $0.041 \pm 0.003$\\
        Scale KS & $\textbf{0.078}~\mathbf{\pm}~\textbf{0.022}$ &
            $\textbf{0.065}~\mathbf{\pm}~\textbf{0.013}$ & $0.056 \pm
            0.009$ & $0.056 \pm 0.005$ & $0.054 \pm 0.008$ & $0.055 \pm 0.004$\\
        Scale KRR & $0.102 \pm 0.033$ & $0.095 \pm 0.100$ & $0.057 \pm
            0.014$ & $0.052 \pm 0.010$ & $0.044 \pm 0.004$ & $0.042 \pm 0.002$\\
        \bottomrule
    \end{tabular}%
    }
    \caption{
    $1$ standard deviation intervals for the mean squared errors of various algorithms when transferring from kin-8fm to kin-8nh.
    The values in bold are the smallest errors for each $n_{ta}$. Only Source KS has much worse performance than other algorithms so we do not show its result here.
    }
    \label{tab:kin_8fm_to_8nh}
\end{table*}


For the first experiment, we vary the size of the target domain to study the effect of $n_{ta}$ relative to $n_{so}$.
We use two datasets from the `kin' family in Delve~\citep{rasmussen1996delve}.
The two datasets we use are `kin-8fm' and `kin-8nh', both with 8 dimensional inputs.
kin-8fm has fairly linear output, and low noise.
kin-8nh on the other hand has non-linear output, and high noise.
We consider the task of transfer learning from kin-8fm to kin-8nh.
In this experiment, We set $n_{so}$ to 320, and vary $n_{ta}$ in $\{10, 20, 40, 80, 160, 320\}$.
Hyper-parameters were picked using grid search with 10-fold cross-validation on the target data (or source domain data when not using the target domain data).

\begin{figure*}[t]
    \centering
    \includegraphics[width=\linewidth]{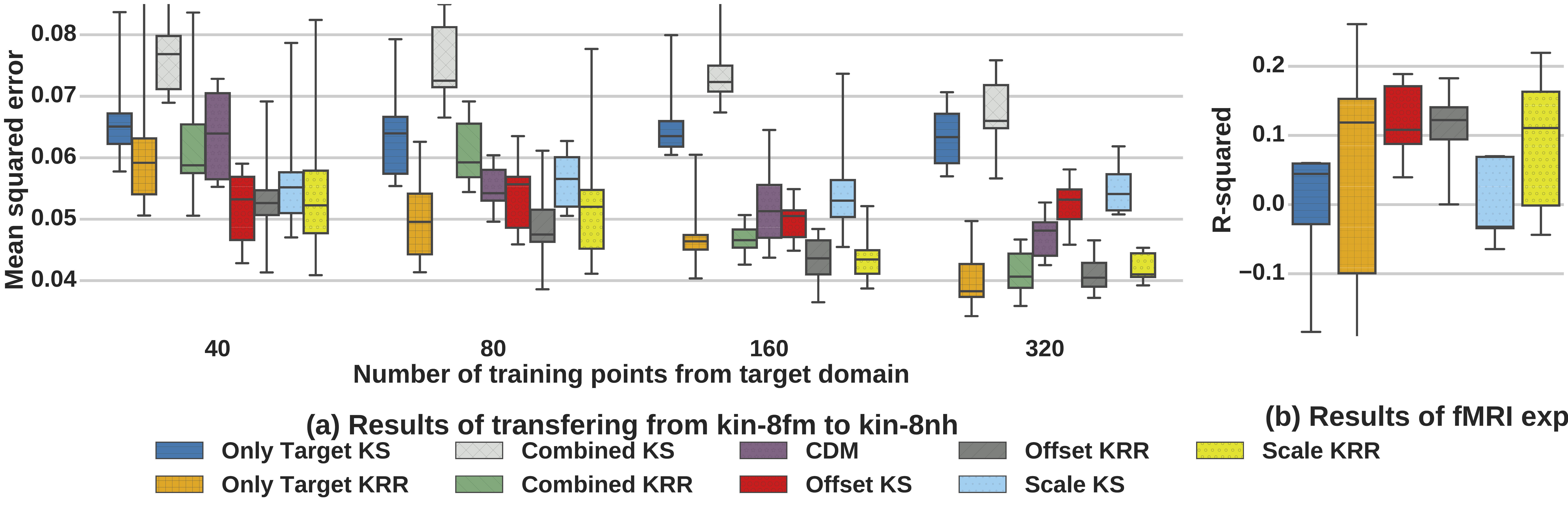}
    \caption{Box plots of experimental results on real datasets. Each box extends from the first to third quartile, and the horizontal lines in the middle are medians. 
    For the robotics data, we report mean squared error (lower the better) and for the fMRI data, we report R-squared (the higher the better).
    For the ease of presentation, we only show results of algorithms with good performances.
    }
    \label{fig:exp_main}
\end{figure*}

Table~\ref{tab:kin_8fm_to_8nh} shows the mean squared errors on the target data.
To better understand the results, we show a box plot of the
mean squared errors for $n_{ta} = 40$ onwards in Figure~\ref{fig:exp_main}(a).
The results for $n_{ta} = 10$ and $n_{ta} = 20$ have high variance, so we do not show them in the plot.
We also omit the results of Only Source KRR because of its poor performance.
We note that our proposed algorithm outperforms other methods across nearly all values of $n_{ta}$ especially when $n_{ta}$ is small.
Only when there are as many points in the target as in the source, does simply training on the target give the best performance.
This is to be expected since the primary purpose in doing transfer learning is to alleviate the problem of lack of data in the target domain.
Though quite comparable, the performance of the scale methods was worse than the offset methods in this experiment.
In general, we would use cross-validation to choose between the two. 

We now consider another real-world dataset where the covariates are fMRI images taken while
subjects perform a Stroop task~\citep{stroop1935studies}.
We use the dataset collected by~\citet{verstynen2014organization} which contains fMRI data of 28 subjects.
A total of 120 trials were presented to each participant 
and fMRI data was collected throughout the trials, and went through a standard post-processing scheme.
The result of this is a feature vector corresponding to each trial that describes the activity of brain regions (voxels), and the goal is to use this to predict the response time.

To frame the problem in the transfer learning setting, we consider as source the data of all but one subject.
The goal is to predict on the remaining subject.
We performed five repetitions for each algorithm by drawing $n_{so} = 300$ data points randomly from the $3000$ points in the source domain.
We used $n_{ta} = 80$ points from the target domain for training and cross-validation;
evaluation was done on the $35$ remaining points in the target domain.
Figure~\ref{fig:exp_main} (b) shows a box plot of the coeffecient of determination values (R-squared)
for the best performing algorithms. R-squared is defined as $1 - SS_{res}/SS_{tot}$ where
$SS_{res}$ is the sum of squared residuals, and $SS_{tot}$ is the total sum of squares. 
Note that R-squared can be negative when predicting on unseen samples -- which were not used to fit the
model -- as in our case. 
When positive, it indicates the proportion of explained variance in the dependent variable (higher the better). 
From the plot, it is clear that Offset KRR and Only Target KRR have the best performances on average and Offset KRR has smaller variance.

Table~\ref{tab:app_fmri} shows the full table of results for the fMRI task.
Using only the source data produces large negative R-squared, and while Only Target KRR does produce a positive
mean R-squared, it comes with a high variance. On the other hand, both Offset methods have low variance, showing
consistent performance. For this particular case, the Scale methods do not perform as well as the Offset methods,
and as has been noted earlier, in general we would use cross validation to select an appropriate transfer function.

\begin{table*}[!t]
    \centering
    \begin{tabular}{@{}llll@{}}
        \toprule
        & Mean & Median & Standard Deviation\\
        \midrule
        Only Target KS & -0.0096 & 0.0444 & \qquad 0.1041\\
        Only Target KRR & 0.1041 & 0.1186 & \qquad 0.2361\\
        Only Source KS & -0.4932 & -0.5366 & \qquad 0.4555\\
        Only Source KRR & -0.8763 & -0.9363 & \qquad 0.6265\\
        Combined KS & -0.7540 & -0.2023 & \qquad 1.5109\\
        Combined KRR & -0.5868 & -0.0691 & \qquad 1.3223\\
        CDM & -3.1183 & -3.4510 & \qquad 2.6473\\
        Offset KS & \textbf{0.1190} & 0.1081 & \qquad \textbf{0.0612}\\
        Offset KRR & 0.1080 & \textbf{0.1221} & \qquad 0.0682\\
        Scale KS & 0.0017 & -0.0321 & \qquad 0.0632\\
        Scale KRR & 0.0897 & 0.1107 & \qquad 0.1104\\
        \bottomrule
    \end{tabular}
    \caption{Mean, median, and standard deviation for the coefficient of determination
        (R-squared) of various algorithms on the fMRI dataset.}
    \label{tab:app_fmri}
\end{table*}

%% file: conclusion.tex
In this paper, we proposed a general transfer learning framework for the HTL regression problem when there is some data available from the target domain.
Theoretical analysis shows it is possible to achieve better statistical rate using transfer learning than standard supervised learning. 

Now we list two future directions and how our results could be further improved.
First, in many real world applications, there is also a large amount of \emph{unlabeled} data from the target domain available.
Combining our proposed framework with previous works for this scenario~\citep{cortes2014domain,huang2006correcting} is a promising direction to pursue.
Second, we only present upper bounds in this paper. 
It is an interesting direction to obtain lower bounds for HTL and other transfer learning scenarios.

%% file: ack.tex
S.S.D. and B.P. were supported by NSF grant IIS1563887 and ARPA-E Terra program. 
A.S. was supported by AFRL grant FA8750-17-2-0212.

%% file: appendix.tex
\section{Proofs}
\subsection{Proof of Theorem~\ref{thm:general_thm}}
\emph{Proof of Theorem~\ref{thm:general_thm}}.\label{pf:general_thm}
The proof just uses assumptions on the transformation function and stability of the training algorithm.
\begin{align}
&\abs{\hat{f}^{ta}\left(x\right)-f^{ta}\left(x\right)}^2 \nonumber \\
= & \abs{G\left(\hat{f}^{so}\left(x\right),\hat{w}_G\left(x\right)\right)-G\left(f^{so}\left(x\right),w_G\left(x\right)\right)}^2 \label{eqn:general_pf_expand_fta}\\
\le &
L^2 \abs{\hat{f}^{so}\left(x\right) - f^{so}\left(x\right)}^2 + L^2 \abs{\hat{w}_G\left(x\right)-w_G\left(x\right)}^2
 \label{eqn:general_pf_G_lipschitz}\\
\le &
L^2 \abs{\hat{f}^{so}\left(x\right) - f^{so}\left(x\right)}^2 +
2L^2 \abs{\hat{w}_G\left(x\right)-\tilde{w}_G\left(x\right)}^2 +
2L^2 \abs{\tilde{w}_G\left(x\right)-w_G\left(x\right)}^2
 \label{eqn:general_pf_G_(a+b)2<a2+b2}\\
\le &
L^2 \abs{\hat{f}^{so}\left(x\right) - f^{so}\left(x\right)}^2 +
2L^2 \left(\sum_{i=1}^{n_{ta}}c_i\left(X_i^{ta}\right)\abs{W_i-\widetilde{W}_i}\right)^2 +
2L^2 \abs{\tilde{w}_G\left(x\right)-w_G\left(x\right)}^2 \label{eqn:general_pf_Awg_stability}
\end{align}
where~(\ref{eqn:general_pf_expand_fta}) is by the requirement of $G$,~(\ref{eqn:general_pf_G_lipschitz}) is by the Lipschitz condition of $G$,~(\ref{eqn:general_pf_G_(a+b)2<a2+b2}) is because $(a-b)^2 \le 2(a-c)^2 + 2(c-b)^2$ and~(\ref{eqn:general_pf_Awg_stability}) is by our stability assumption of $\mathcal{A}_{w_G}$.
Now, we are left bounding $\left(\sum_{i=1}^{n_{ta}}c_i\abs{W_i-\widetilde{W}_i}\right)^2 $.
Notice that by the assumption of $H_G$, \begin{align}
\abs{W_i - \widetilde{W}_i}  = \abs{H_G\left(\hat{f}^{so}\left(X_i^{ta}\right),\left(Y_i^{ta}\right)\right)
-H_G\left(f^{so}\left(X_i^{ta}\right),Y_i^{ta}\right)
} \le L\abs{\hat{f}^{so}\left(X_i^{ta}\right) - f^{so}\left(X_i^{ta}\right)} \label{eqn:general_of_G_inverse_lipschitz}
\end{align}
Plugging~(\ref{eqn:general_of_G_inverse_lipschitz}) into~(\ref{eqn:general_pf_Awg_stability}), we obtain our desired result.
\qed

\subsection{Proof of Theorem~\ref{thm:kernel_smoothing_global}}
For simplicity, let $K_h (\cdot) = K(\cdot/h)$ and define the expected regression estimate $\tilde{f} = \sum_{i=1}^{n}w_if(X_i)$.
To prove Theorem~\ref{thm:kernel_smoothing_global}, we first give some standard supporting lemmas for kernel smoothing.
\begin{lemma}[Lemma 1 of~\citep{kpotufe2013adaptivity}]\label{thm:bias}
Under the same assumptions in Theorem~\ref{thm:kernel_smoothing_global}, for all $x$ with $\norm{x}_2 \le \triangle_X$, if $f$ is $(\lambda,\alpha)$ \holder,
then, for any $h >0$, we have $|\tilde{f}(x)-f(x)|^2 \le \lambda^2h^{2\alpha}$.
\end{lemma}
\begin{lemma}[Corollary of Lemma 3 and Lemma 7 of~\citep{kpotufe2013adaptivity}]\label{thm:variance}
Under the same assumptions in Theorem~\ref{thm:kernel_smoothing_global}, let $0 < \delta < 1/6$, for all $x:\norm{x}_2 \le \triangle_X$ and $h >0$, with probability at least $1-\delta$, we have
\begin{align*}
|\hat{f}(x) - \tilde{f}(x)|^2 = O\left(\frac{\log\left(1/\delta\right)}{nh^d}\right).
\end{align*}
\end{lemma}

\emph{Proof of Theorem~\ref{thm:kernel_smoothing_global}}.
we prove Theorem~\ref{thm:kernel_smoothing_global} by bounding each corresponding term in Theorem~\ref{thm:general_thm}.
First, by Lemma~\ref{thm:bias} and Lemma~\ref{thm:variance}, we have for all $x$, with probability at least $1-\delta$  \begin{align*}
 \abs{\hat{f}^{so}(x)-f^{so}(x)}^2 = O\left(h_{so}^{2\alpha_{so}} + \frac{\log\left(1/\delta\right)}{n_{so} h_{so}^{d}}\right).
\end{align*}
Specifically, for $X_1^{ta},\ldots,X_{n_{ta}}^{ta}$, we have
\begin{align}
 \max_{i=1,\cdots,n_{ta}}\abs{\hat{f}^{so}(X_i^{ta})-f^{so}(X_i^{ta})}^2 = O\left(h_{so}^{2\alpha_{so}} + \frac{\log\left(1/\delta\right)}{n_{so} h_{so}^{d}}\right). \label{eqn:kernel_max}
\end{align}
Next, according to Assumption~\ref{assum:G_lipschitz} and~\ref{assum:w_bounded}, $H_G$ is bounded and unbiased and $w_G$ is bounded, we can view $\left\{\left(X_i^{ta}, \widetilde{W}_i\right)\right\}_{i=1}^{n_{ta}}$ a training set for function $w_G$ that $\widetilde{W}_i = w_G\left(X_i^{ta}\right) + \epsilon_{w_G}$ where $\mathbf{E}\left[\epsilon_{w_G}\right] = 0$ and $\abs{\epsilon_{w_G}} \le 2B$.
Based on this observation, using Lemma~\ref{thm:bias} and Lemma~\ref{thm:variance} again, for all $x: \norm{x}_2 \le \triangle_X$, we have with probability at least $1-\delta$ \begin{align*}
\left|\tilde{w}_G\left(x\right)-w_G\left(x\right)\right|^2
=  O\left(h_{w_G}^{2\alpha_{w_G}} + \frac{\log\left(1/\delta\right)}{n_{ta}h_{w_G}^d}\right).
\end{align*}
Now we are left bounding $\norm{\mathcal{A}_{w_G}\left(\mathcal{T}\right) - \mathcal{A}_{w_G}\left(\widetilde{\mathcal{T}}\right)}_{\infty}$.
Notice that for $\mathcal{T}$, $\mathcal{\tilde{T}}$ in Theorem~\ref{thm:general_thm}, and for all $x: \norm{x}_2 \le \triangle_X$:
\begin{align*}
\abs{\mathcal{A}_{w_G}\left(\mathcal{T}\right)\left(x\right) - \mathcal{A}_{w_G}\left(\widetilde{\mathcal{T}}\right)\left(x\right)}
&= \frac{\sum_{i=1}^{n_{ta}}K_h\left(\norm{x-X_i^{ta}}_2\right)\left(W_i - \widetilde{W}_i\right)}{\sum_{i=1}^{n_{ta}}K_h\left(\norm{x-X_i^{ta}}_2\right)} \\
&= \frac{\sum_{i=1}^{n_{ta}}K_h\left(\norm{x-X_i^{ta}}_2\right)\left|W_i - \widetilde{W}_i\right|}{\sum_{i=1}^{n_{ta}}K_h\left(\norm{x-X_i^{ta}}_2\right)} \\
& \triangleq \sum_{i=1}^{n_{ta}}c_i\left|W_i-\widetilde{W}_i\right|
\end{align*}
for $c_i = \frac{K_h\left(\norm{x-X_i^{ta}}_2\right)}{\sum_{i=1}^{n_{ta}}K_h\left(\norm{x-X_i^{ta}}_2\right)}$.
Now according to Theorem~\ref{thm:general_thm}, we only need to bound $\left(\sum_{i=1}^{n_{ta}}c_i\left|\hat{f}^{so}\left(X_i^{ta}\right)-f^{so}\left(X_i^{ta}\right)\right|\right)^2$.
With probability at least $1-\delta$, we have:
 \begin{align}
\left(\sum_{i=1}^{n_{ta}}c_i\left|\hat{f}^{so}\left(X_i^{ta}\right)-f^{so}\left(X_i^{ta}\right)\right|\right)^2
& \le \left(\sum_{i=1}^{n_{ta}}c_i\right)^2   \left(\max_{i=1,\cdots,n_{ta}}\left|\hat{f}^{so}\left(X_i^{ta}\right)-f^{so}\left(X_i^{ta}\right)\right|^2\right) \label{eqn:pf_smoothing_max_bigger_than_all}\\
& = \max_{i=1,\cdots,n_{ta}}\left|\hat{f}^{so}\left(X_i^{ta}\right)-f^{so}\left(X_i^{ta}\right)\right|^2 \label{eqn:pf_smoothing_sum_of_ci}\\
& = O \left(h_{so}^{2\alpha_{so}} + \frac{\log\left(n_{ta}/\delta\right)}{n_{so} h_{so}^{d}} \right) \label{eqn:pf_smoothing_bias_var},
\end{align} where~\eqref{eqn:pf_smoothing_max_bigger_than_all} is because maximum is bigger than other terms,~\eqref{eqn:pf_smoothing_sum_of_ci} is because $\sum_{i=1}^{n_{ta}}c_i = 1$ by definition,
and~\eqref{eqn:pf_smoothing_bias_var} is by~\eqref{eqn:kernel_max}.
Putting these all together, using Theorem~\ref{thm:general_thm} and choosing the bandwidth according to Theorem~\ref{thm:kernel_smoothing_global}, we can show for all $x: \norm{x}_2 \le \triangle_X$ \[
\abs{f^{ta}(x)-\hat{f}^{ta}(x)}^2 = O\left( n_{so}^{\frac{-2\alpha_{so}}{2\alpha_{so}+d}} + n_{ta}^{\frac{-2\alpha_{w_G}}{2\alpha_{w_G}+d}}\right)\log\left(\frac{1}{\delta}\right).
\]
Now integrate with respect to $P_{X^{ta}}$ we obtain our desired result.
\qed

\subsection{Proof of Theorem~\ref{thm:RKHS_regression}}
The proof strategy is similar to that of Theorem~\ref{thm:kernel_smoothing_global}.
Using Theorem~\ref{thm:general_thm} we have \begin{align*}
\mathbf{E}\left[\left|\hat{f}^{ta}(X)-f^{ta}(X)\right|^2\right]  =  & O\left(\mathbf{E}\left[
\left|\hat{f}^{so}\left(X\right)-f^{so}\left(X\right)\right|^2 + \left|\widetilde{w}_G\left(X\right) - w_G\left(X\right)\right|^2
+ \right. \right.\\
& \left.  \left. \left(\sum_{i=1}^{n_{ta}}c_i\left(X_i^{ta}\right)\left|\hat{f}^{so}\left(X_i^{ta}\right)-f^{so}\left(X_i^{ta}\right)\right|\right)^2\right]
\right).
\end{align*}
where the expectation is taken over $P_{x^{ta}}$ and $\mathcal{T}^{ta}$.
Now we bound three terms on the right hand side separately.
By Corollary 3 of~\cite{steinwart2009optimal}, we have with probability at least $1-\delta$\begin{align}
\mathbf{E}\left[\abs{\hat{f}^{so}\left(X\right) - f^{so}\left(X\right)}^2\right] = O\left(\lambda_{so}^{\beta_{so}} + \frac{\log\left(1/\delta\right)}{\lambda_{so}^p n_{so}}\right), \label{eqn:rkhs_bound}
\end{align} where expectation is taken over $P_x^{ta}$.
Taking union bound over $X_1^{ta},\ldots,X_{n_{ta}}^{ta}$, we have
\begin{align}
 \max_{i=1,\cdots,n_{ta}}\mathbf{E}\left[\abs{\hat{f}^{so}(X_i^{ta})-f^{so}(X_i^{ta})}^2\right] = O\left(\lambda_{so}^{\beta_{so}} + \frac{\log\left(n_{ta}/\delta\right)}{\lambda_{so}^p n_{so}}\right). \label{eqn:rkhs_max}
\end{align}
where the expectation is taken over $\mathcal{T}^{ta}$.
Next, using the exactly same argument as in the Theorem~\ref{thm:kernel_smoothing_global}, we can view $\left\{\left(X_i^{ta}, \widetilde{W}_i\right)\right\}_{i=1}^{n_{ta}}$ a training set for function $w_G$ that $\widetilde{W}_i = w_G\left(X_i^{ta}\right) + \epsilon_{w_G}$ as $\widetilde{W}_i = w_G\left(X_i^{ta}\right) + \epsilon_{w_G}$ where $\mathbf{E}\left[\epsilon_{w_G}\right] = 0$ and $\abs{\epsilon_{w_G}} \le 2B$.
Thus applying Corollary 3 of~\cite{steinwart2009optimal} again, we have with probability at least $1-\delta$
\begin{align*}
\mathbf{E}\left[\left|\tilde{w}_G\left(X\right)-w_G\left(X\right)\right|^2\right] = O\left(\lambda_{w_G}^{\beta_{w_G}} + \frac{\log\left(1/\delta\right)}{\lambda_{w_G}^pn_{ta}}\right).
\end{align*}
where expectation is taken over $P_{x^{ta}}$.
Now we analyze the stability of KRR.
We use $\Phi\left(x\right)$ to denotes the feature map corresponding with the given kernel $K$ so $K(x,y) = \Phi\left(x\right)^\top \Phi\left(y\right)$.
Also for simplicity, we denote \[
\mat{\Phi}_{ta} = \begin{pmatrix}
\Phi\left(x_1^{ta}\right)\mid \cdots \mid\Phi\left(x_{n_{ta}}^{ta}\right)
\end{pmatrix}
\]
the feature matrix of target domain data.
With these notations, we can write
\begin{align*}
& \abs{\mathcal{A}_{w_G}\left(\mathcal{T}^{w_G}\right)\left(x\right) - \mathcal{A}_{w_G}\left(\mathcal{\widetilde{T}}^{w_G}\right)\left(x\right)} \\
= & \abs{\begin{pmatrix}
W_1 - \widetilde{W}_1 \\
\cdots \\
W_{n_{ta}} - \widetilde{W}_{n_{ta}}
\end{pmatrix}^\top \left(\mat{\Phi}_{ta}^\top \mat{\Phi} + n_{ta}\lambda_{w_G}\mat{I}\right)^{-1}\mat{\Phi}_{ta}^\top \Phi\left(x\right)} \\
= & \abs{
\left(
\mat{\Phi}_{ta}
\begin{pmatrix}
W_1 - \widetilde{W}_1 \\
\cdots \\
W_{n_{ta}} - \widetilde{W}_{n_{ta}}
\end{pmatrix}
\right)^\top \left(\mat{\Phi}_{ta}\mat{\Phi}_{ta}^\top + n_{ta}\lambda_{w_G}\mat{I}\right)^{-1}\Phi\left(x\right)
} \\
\le &  \norm{
\begin{pmatrix}
k^{1/2}\abs{W_1-\widetilde{W}_1}\\
\cdots \\
k^{1/2}\abs{W_{n_{ta}}-\widetilde{W}_{n_{ta}}}_2
\end{pmatrix}
}_2 \norm{\left(\mat{\Phi}_{ta}\mat{\Phi}_{ta}^\top + n_{ta}\lambda_{w_G}\mat{I}\right)^{-1}}_{op} k^{1/2} \\
\le &   \sum_{i=1}^{n_{ta}}\frac{k}{n_{ta}\lambda_{w_G}}\abs{W_i-\widetilde{W}_i} \\
\triangleq &
\sum_{i=1}^{n_{ta}}c_i\abs{W_i-\widetilde{W}_i}.
\end{align*}
The second equality we used the identity that $
\left(\mat{\Phi}^\top\mat{\Phi} + \lambda \mat{I}\right)^{-1}\mat{\Phi}^\top = \mat{\Phi}^\top\left(\mat{\Phi}\mat{\Phi}^\top+\lambda\mat{I}\right)^{-1}
$ for any $\mat{\Phi}$ and $\lambda$.
The first inequality we used sub-multiplicity of operator norm and the assumption $\norm{\Phi\left(x\right)}_{\mathcal{H}} \le k^{1/2}$.
The second inequality we used the fact the lower bound of least eigenvalue of $\left(\mat{\Phi}_{ta}\mat{\Phi}_{ta}^\top + n_{ta}\lambda_{w_G}\mat{I}\right)$ is $n_{ta}\lambda_{w_G}$.
Therefore, applying Cauchy-Schwartz inequality and using the bound in~\eqref{eqn:rkhs_max}, we have with probability at least $1-\delta$,
\begin{align*}
\mathbf{E}\left[\left(\sum_{i=1}^{n_{ta}}c_i\left|\hat{f}^{so}\left(X_i^{ta}\right)-f^{so}\left(X_i^{ta}\right)\right|\right)^2 \right]
& \le\left(\sum_{i=1}^{n_{ta}}c_i^2\right) \cdot  \left(\sum_{i=1}^{n_{ta}}\left|\hat{f}^{so}\left(X_i^{ta}\right)-f^{so}\left(X_i^{ta}\right)\right|^2\right) \\
& = \sum_{i=1}^{n_{ta}}\frac{k^2}{n_{ta}^2\lambda_{w_G}^2}\cdot \mathbf{E}\left[\sum_{i=1}^{n_{ta}}\left|\hat{f}^{so}\left(X_i^{ta}\right)-f^{so}\left(X_i^{ta}\right)\right|^2\right] \\
& \le \frac{k^2}{\lambda_{w_G}^2}\cdot \max_{i=1,\ldots,n_{ta}}\mathbf{E}\left[\left|\hat{f}^{so}\left(X_i^{ta}\right)-f^{so}\left(X_i^{ta}\right)\right|^2\right] \\
& = O\left(\frac{k^2}{\lambda_{w_G}^2}\left(\lambda_{so}^{\beta_{so}} + \frac{\log\left(n_{ta}/\delta\right)}{\lambda_{so}^pn_{so}}\right)\right).
\end{align*}
Now putting these all together and choosing $\lambda_{so}$ and $\lambda_{w_G}$ according to Theorem~\ref{thm:RKHS_regression}, we obtain the desired result.
\qed


\subsection{Proof of Theorem~\ref{thm:best_in_class}}
We first prove a general theorem for cross-validation.
This is a standard result for cross-validation and we include the proof for completeness.
\begin{theorem}\label{thm:safe}
Let $\Theta$ be the set of all hypotheses and $\hat{\theta} = \argmin_{\theta \in \Theta} \sum_{i=1}^{n_{val}}\left(\hat{f}^{ta}_{\theta}\left(X_i^{val}\right) - Y_i^{val}\right)^2$ the estimator that minimizes error on the cross-validation set.
Then with probability at least $1-\delta$:
\begin{align*}
\mathbf{E}\left[R\left(\hat{f}^{ta}_{\hat{\theta}}\right)\right] - R\left(f^{ta}\right) = O\left(
\mathbf{E}\left[R\left(\hat{f}^{ta}_{\theta_\star}\right)\right]-R\left(f^{ta}\right) + \frac{\log\frac{\left|\Theta\right|}{\delta}}{n_{val}}\right),
\end{align*} where $\theta^* = \argmin_{\theta \in \Theta}R\left(\hat{f}_{\theta}\right)$ and the expectation is taken over $\mathcal{T}^{so}$ and $\mathcal{T}^{ta}$.
\end{theorem}

To prove of Theorem~\ref{thm:safe}, we use the following type of Bernstein's inequality~\citep{craig1933tchebychef}:
\begin{lemma}\label{thm:bernstein}
Let $X_1,\ldots,X_n$ be random variables and suppose that for $k \ge 3$ :
\begin{align*}
\mathbf{E}[\left|X_i - \mathbf{E}[X_i]\right|^k] \le \frac{\mathbf{Var}[X_i]}{2}k!r^{k-2},
\end{align*} for some $r > 0$. Then with probability $ > 1-\delta$:\begin{align*}
\frac{1}{n}\sum_{i=1}^{n}\left(X_i - \mathbf{E}[X_i]\right) \le \frac{\log(1/\delta)}{nt} + \frac{t\mathbf{Var}[X_i]}{2(1-c)},
\end{align*} for $0 \le tr \le c < 1$.
\end{lemma}
\emph{Proof of Theorem~\ref{thm:safe}:}
For a given $\theta \in \Theta$, we obtain a corresponding estimated regression function $\hat{f}_\theta$.
Define $U_i^\theta \triangleq -\left(Y^{val}_i - \hat{f}^{ta}_{\theta}(X^{val}_i)\right)^2 + (Y_i - f^{ta}(X_i^{val}))^2$.
Compute the expectation:
\begin{align*}
\mathbf{E}\left[U_i^\theta\right]
= & -\mathbf{E}\left[
-2Y_i^{val}\hat{f}_{\theta}^{ta}\left(X_i^{val}\right) + \hat{f}^{ta}_{\theta}\left(X_i^{val}\right)^2 + 2Y_i^{val}f^{ta}\left(X_i^{val}\right) - f^{ta}\left(X_i^{val}\right)^2
\right] \\
= & -\mathbf{E}\left[\left(\hat{f}^{ta}_{\theta}\left(X_i^{val}\right)-f^{ta}\left(X_i^{val}\right)\right)^2\right]\\
= & R\left(f^{ta}\right) - R\left(\hat{f}_{\theta}^{ta}\right).
\end{align*}
Also, by definition, it is easy to see\begin{align*}
\frac{1}{n_{val}}\sum_{i=1}^{n_{val}}U^\theta_i = \hat{R}\left(f\right) - \hat{R}\left(\hat{f}^{ta}_{\theta}\right).
\end{align*}
In order to apply Bernstein's inequality, we must first bound  the variance of $U_i^\theta$:
\begin{align*}
\mathbf{var}\left[U_i^\theta\right]& \le \mathbf{E}\left[(U_i^\theta)^2\right] \\
& = \mathbf{E}\left[
\left(-\left(Y_i^{val}-\hat{f}_\theta^{val}\left(X_i^{val}\right)\right)^2
+\left(Y_i^{val} - f^{ta}\left(X_i^{val}\right)\right)^2
\right)^2
\right] \\
& = \mathbf{E}\left[
\left(f^{ta}\left(X_i\right)-\hat{f}^{ta}_\theta\right)^4 + 4\epsilon_i\left(f^{ta}\left(X_i^{vak}\right)-\hat{f}^{ta}_\theta\left(X_i^{val}\right)\right)^3 + 4\epsilon_i^2 \left(f^{ta}\left(X_i^{val}\right) - \hat{f}^{ta}_\theta\left(X_i^{val}\right)\right)^2
\right] \\
& \le -4\triangle_Y^2 \mathbf{E}\left[U_i\right]
\end{align*} where in the last inequality we used the domain of $Y$ is bounded.
Since $\mathbf{U}_i$ is a sum of bounded random variables, the moment condition is satisfied with $r = 4\triangle_Y^2$.
Now apply Craig-Bernstein inequality to $U_i^\theta$s, with probability at least $1-\delta$:
\begin{align*}
\frac{1}{n_{val}}\sum_{i=1}^{n_{val}}U_i^\theta - \mathbf{E}\left[U_i^\theta\right]
 \le \frac{\log(1/\delta)}{n_{val} t} + \frac{-2t\triangle_Y^2\mathbf{E}\left[U_i^\theta\right]}{1-c}.
\end{align*}
We need to ensure that $c < 1$. To do this, let $c = tr = 4t\triangle_Y^2$ and let $t < \frac{1}{6\triangle_Y^2}$, then it is easy to see that $c < 1$.
For simplicity, define $a = \frac{2t\triangle_Y^2}{1-c} < 1$.
Now grouping terms we get:
\begin{align*}
\left(1-a\right)(-\mathbf{E}\left[U_i^\theta\right]) + \frac{1}{n_{val}}\sum_{i=1}^{n_{val}}U_i^\theta & \le \frac{\log\left(1/\delta\right)}{n_{val}t} \\
\left(1-a\right)\left(R\left(\hat{f}^{ta}\right) - R\left(f\right)\right) - \left(\hat{R}\left(f^{ta}_\theta\right) - \hat{R}\left(f\right)\right) & \le \frac{\log\left(1/\delta\right)}{n_{val}t}\\
R\left(\hat{f}^{ta}_\theta\right) - R\left(f^{ta}\right) & \le
\frac{1}{1-a} \left(\hat{R}\left(\hat{f}_\theta\right)- \hat{R}\left(f^{ta}\right) + \frac{\log\left(1/\delta\right)}{n_{val}t}\right).\\
\end{align*}
Take union bound over $\Theta$, and consider $\hat{f}_{\hat{\theta}}$:
\begin{align*}
	R\left(\hat{f}^{ta}_{\hat{\theta}}\right) - R\left(f^{ta}\right) & \le
	\frac{1}{1-a} \left(\hat{R}\left(\hat{f}^{ta}_{\hat{\theta}}\right)-\hat{R}\left(f^{ta}\right)+ \frac{\log\left(\left|\Theta\right|/\delta\right)}{n_{val}t}\right) .
\end{align*}
Now, recall that $\hat{f}^{ta}_{\hat{\theta}}$ is the minimizer for $\hat{R}$ among all estimators induced by $\Theta$, we have
\begin{align*}
R\left(\hat{f}^{ta}_{\hat{\theta}}\right) - R\left(f^{ta}\right) & \le
\frac{1}{1-a} \left(\hat{R}\left(\hat{f}^{ta}_{\theta_\star}\right)-\hat{R}\left(f^{ta}\right) + \frac{\log\left(\left|\Theta\right|/\delta\right)}{n_{val}t}\right).
\end{align*}
Now taking expectation over $\mathcal{T}^{val}$ then over $\mathcal{T}^{so}$ and $\mathcal{T}^{ta}$ we obtain the desired result.
\qed

Now we are ready to prove Theorem~\ref{thm:best_in_class}.
Since $\overline{\mathcal{G}}$ is an $\epsilon$-cover of $\mathcal{G}$, there exists $G' \in \overline{\mathcal{G}}$ such that $\norm{G'-G^\star}_\infty \le \epsilon$.
For any $x$, \begin{align}
 &\abs{f^{ta}\left(x\right) - \hat{f}^{ta}_{G'}\left(x\right)} \nonumber\\ =&\abs{G^\star\left(f^{so}\left(x\right),w_{G^\star}\left(x\right)\right) - G'\left(\hat{f}^{so}\left(x\right),\hat{w}_{G'}\left(x\right)\right)} \nonumber \\
 \le &  \abs{G^\star\left(f^{so}\left(x\right),w_{G^\star}\left(x\right)\right) - G^\star\left(\hat{f}^{so}\left(x\right),\hat{w}_{G^\star}\left(x\right)\right)} + \abs{G^\star\left(\hat{f}^{so}\left(x\right),\hat{w}_{G^\star}\left(x\right)\right) - G'\left(\hat{f}^{so}\left(x\right),\hat{w}_{G^\star}\left(x\right)\right)}
 \nonumber \\
 &+ \abs{G'\left(\hat{f}^{so}\left(x\right),\hat{w}_{G^\star}\left(x\right)\right) - G'\left(\hat{f}^{so}\left(x\right),\hat{w}_{G'}\left(x\right)\right)} \label{eqn:best_class_3terms}
\end{align} where $\hat{w}_{G^\star} = \mathcal{A}_{w_G}\left(\left\{X_i^{ta}, W_i^\star \right\}\right)$ and $W_i^\star = H_{G'}\left(\hat{f}^{so}\left(X_i^\star\right), Y_i^\star\right) + w_{G^\star}\left(X_i^{ta}\right) - w_{G'}\left(X_i^{ta}\right)$, i.e. an un-biased estimated of $w_{G^\star}\left(X_i^\star\right)$.
We can bound three terms in~\eqref{eqn:best_class_3terms} separately.
The first term is just the difference between estimator based on $G^\star$ and the true $f^{ta}$, so after taking expectation it becomes the excess risk of $\hat{f}^{ta}_{G^\star}$.
By our construction of $\epsilon$-cover of $\mathcal{G}$, the second term is smaller than $\epsilon$.
For the third term, notice that by Lipschitz assumption on $G$s and our assumptions on $G$s in $\mathcal{G}$ in the theorem~\ref{thm:best_in_class}, we have:\begin{align*}
&\abs{G'\left(\hat{f}^{so}\left(x\right),\hat{w}_{G^\star}\left(x\right)\right) - G'\left(\hat{f}^{so}\left(x\right),\hat{w}_{G'}\left(x\right)\right)}  \\
\le & L\left(\abs{\hat{w}_{G^\star}\left(x\right)-\hat{w}_{G'}\left(x\right)}\right)\\
\le & L^2 \sum_{i=1}^{n_{ta}}c_i\norm{G^\star - G'}_\infty\\
=  &O\left(\sum_{i=1}^{n_{ta}}c_i\epsilon\right).
\end{align*}
Now we have shown $R\left(\hat{f}^{ta}_{G'}\right) - R\left(f^{ta}\right) = O\left(R\left(\hat{f}^{ta}_{G^\star}\right) - R\left(f^{ta}\right)\right)$.
Let $\overline{G}_\star = \argmin_{G\in \mathcal{\overline{G}}}R\left(\hat{f}_{G}\right)$, the best transformation function in $\mathcal{\overline{G}}$.
By the optimality of $\overline{G}_\star$, we have $R\left(\hat{f}^{ta}_{\overline{G}_\star}\right) - R\left(f^{ta}\right) = O\left(R\left(\hat{f}^{ta}_{G^\star}\right) - R\left(f^{ta}\right)\right)$.
Applying Theorem~\ref{thm:safe} with our assumptions on $\epsilon$ and $n_{val}$ we know
$R\left(\hat{f}^{ta}_{\overline{G}^\star}\right) - R\left(f^{ta}\right) = O\left(R\left(\hat{f}^{ta}_{\overline{G}_\star}\right) - R\left(f^{ta}\right)\right)$.
Combing these facts we have $R\left(\hat{f}^{ta}_{\overline{G}^\star}\right) - R\left(f^{ta}\right) = O\left(R\left(\hat{f}^{ta}_{G_\star}\right) - R\left(f^{ta}\right)\right)$.

\section{Regression Calibration for Measurement Error Problem}\label{sec:measure_err}
Given, $f^{so}$, in this section we provide a standard technique to obtain an unbiased estimate of $w_G\left(X_i^{ta}\right)$s.
Since we assume
\[
Y^{ta} = f^{ta}\left(X^{ta}\right) + \epsilon^{ta},
\]
the measurement error model corresponds to \emph{classical error model} in~\cite{carroll2006measurement}.
Regression calibration is a widely used and reasonably well investigated method for measurement error problem.
The algorithm is as follows (we have adapted the general algorithm to our HTL problem):
\begin{itemize}
\item Compute an estimate of $f^{ta}\left(X_i^{ta}\right)$: $\tilde{f}^{ta}\left(X_i^{ta}\right)$.
Note that directly using $Y_i^{ta}$ is one of the option for $\tilde{f}^{ta}\left(X_i^{ta}\right)$.
\item Compute $G_{f^{so}\left(X_i^{ta}\right)}^{-1}\left(\tilde{f}^{ta}\left(X_i^{ta}\right) \right)$.
\item Calibrate our previous computed value by applying some function $F$: \[
\widetilde{W}_i = F\left(G_{f^{so}\left(X_i^{ta}\right)}^{-1}\left(\tilde{f}^{ta}\left(X_i^{ta}\right)  \right)\right)
\] where $F$ depends on $G$ and the specific distribution on noise.
\end{itemize}
Now we consider the loglinear mean model as a concrete example.
Suppose
\[
G\left(f^{so}\left(x\right),w_G\left(x\right)\right) = \beta f^{so}\left(x\right)\log\left(w_G\left(x\right)\right)
\] where $\beta$ is some constant.
Further, we assume $\epsilon^{ta} \sim \mathcal{N}\left(0, \sigma^2\right) $
Now we apply the regression calibration algorithm.
\begin{itemize}
\item First we choose $Y_i^{ta}$ as our estimate for $\tilde{f}^{ta}\left(X_i^{ta}\right)$.
\item Second, by our choice of $G$:\begin{align*}
G^{-1}_{f^{so}\left(X_i^{ta}\right)}\left(Y_i^{ta}\right) = \exp\left(\frac{Y_i^{ta}}{\beta f^{so}\left(X_i^{ta}\right)}\right)
\end{align*}
\item Last, for our choice of $G$ and assumption of $\epsilon^{ta}$, the corresponding $F$ and final estimate of $w_G\left(X_i^{ta}\right)$ is\begin{align*}
\widetilde{W}_i & = F\left(G_{f^{so}\left(X_i^{ta}\right)}^{-1}\left(\tilde{f}^{ta}\left(X_i^{ta}\right)  \right)\right) \\
& = \exp\left(\log\left(G_{f^{so}\left(X_i^{ta}\right)}^{-1}\left(\tilde{f}^{ta}\left(X_i^{ta}\right)\right) \right) + \sigma^2\left(f^{so}\left(X_i^{ta}\right)\right)^2\right)\\
& = \exp\left(\frac{Y_i^{ta}}{\beta f^{so}\left(X_i^{ta}\right)} + \sigma^2\left(f^{so}\left(X_i^{ta}\right)\right)^2\right).
\end{align*}
\end{itemize}
The estimator for $w_G\left(X_i^{ta}\right)$ depends on some distribution specific parameters which may be unknown, like $\sigma^2$ in the previous example.
In such cases, we may replace these parameters by our estimates.
For example, in the previous Gaussian noise case, suppose for each $X_i^{ta}$, we have multiple observations $\left\{Y_{ij}\right\}_{j=1}^{n_i}$.
Then we can estimate $\sigma^2$ by \[
\hat{\sigma}^2 = \frac{\sum_{i=1}^{n_{ta}}\sum_{j=1}^{n_i}\left(Y_{ji}^{ta}-\bar{Y}_i^{ta}\right)^2}{\sum_{i=1}^{n_{ta}}\left(n_i-1\right)}
\]
where $\bar{Y}_i^{ta} = \frac{\sum_{j=1}^{n_i}Y_{ij}}{n_i}$.\\
Here we only provide one method for measurement error problem.
There are other techniques such simulation extrapolation and likelihood method which may be also applicable in many situations.
The choice of method depends on specific transformation $G$ and assumptions on the distribution of the noise.
Again, interested readers are referred to~\cite{carroll2006measurement} for details.

\section{Additional Experimental Results}

\subsection{Synthetic data}\label{sec:syn}
This section gives details of the synthetic data.
For both experiments, we use $n_{so} = 10000$ samples from the source domain, and
$n_{ta} = 100$ samples from the target domain. We put Gaussian noise on the labels:
$\epsilon^{so}\sim \mathcal{N}\left(0,0.01\right)$, $\epsilon^{ta} \sim
\mathcal{N}\left(0,0.01\right)$; and we use KS with a gaussian
kernel for estimating $f^{so}$ and $w_G$.

Figure~\ref{fig:doppler_plus} shows the offset example in Section~\ref{sec:hypo}, where we consider
\[
f^{so}(x) = \sqrt{x\left(1-x\right)}\sin\left(\frac{2.1\pi}{x+0.05}\right),
f^{ta}(x) = f^{so}(x) + x.
\]
We used the transformation function $G(a,b) = a+b$.
The bandwidths of the kernels were chosen by cross validation.
For estimating $f^{so}$, the chosen bandwidth is $h_{so} = 10^{-8}$, and for estimating $w_G$, the chosen value is $h_{w_{G}} = 10^{-5}$.
Figure~\ref{fig:doppler_mult} shows the scale example in Section~\ref{sec:hypo}, where we consider the same source regression function and $f^{ta}(x) = 5f^{so}(x)$.
We tested the transformation function $G(a,b) = ab$.
Bandwidth parameters were again chosen by cross validation: $h_{so} = 10^{-7}$ for
estimating $f^{so}$, and $h_{w_{G}} = 5\times 10^{-4}$ for estimating $w_{G}$.
The plots show that by using our proposed transfer learning framework with an
appropriate transformation function, we can estimate the target regression function
better, especially in regions where $f^{ta}$ is not smooth.

\subsection{Transferring from kin-8nh to kin-8fm}\label{subsec:revtrans}
Now we briefly discuss the results of the second transfer task with the
robotic arm data described in Section~\ref{sec:exp}. The source domain
is kin-8nh and the target domain is kin-8fm. The results are
shown in Table~\ref{tab:kin_8nh_to_8fm}.
Here we see the effects of trying to transfer to an ``easy'' domain.
We do not gain any advantage by using the transfer algorithm, except for the smallest value of
$n_{ta}$ - even here the gain is minimal. However, it should be noted that
using transfer learning does not negatively affect performance. And we point out
that in a dataset where the smoothness conditions are unknown, we would use
cross-validation to decide whether or not to use the source data.

\begin{table*}[!t]
    \centering
    \resizebox{\columnwidth}{!}{%
    \begin{tabular}{@{}lllllll@{}}
        \toprule
        & $n_{ta} = 10$ & $n_{ta} = 20$ & $n_{ta} = 40$ & $n_{ta} = 80$ &
            $n_{ta} = 160$ & $n_{ta} = 320$\\
        \midrule
        Only Target KS & $0.005 \pm 0.001$ & $0.003 \pm 0.001$ & $0.003
            \pm 0.001$ & $0.003 \pm 0.000$ & $0.002 \pm 0.000$ & $0.002 \pm 0.000$\\
        Only Target KRR & $\textbf{0.001}~\mathbf{\pm}~\textbf{0.001}$ &
            $\textbf{0.001}~\mathbf{\pm}~\textbf{0.000}$
            & $\textbf{0.000}~\mathbf{\pm}~\textbf{0.000}$
            & $\textbf{0.000}~\mathbf{\pm}~\textbf{0.000}$
            & $\textbf{0.000}~\mathbf{\pm}~\textbf{0.000}$
            & $\textbf{0.000}~\mathbf{\pm}~\textbf{0.000}$\\
        Only Source KS & $0.031 \pm 0.012$ & $0.031 \pm 0.012$ & $0.031
            \pm 0.012$ & $0.031 \pm 0.012$ & $0.031 \pm 0.012$ & $0.031 \pm 0.012$\\
        Only Source KRR & $0.016 \pm 0.013$ & $0.016 \pm 0.013$ & $0.016
            \pm 0.013$ & $0.016 \pm 0.013$ & $0.016 \pm 0.013$ & $0.016 \pm 0.013$\\
        Combined KS & $0.023 \pm 0.017$ & $0.029 \pm 0.011$ & $0.017 \pm
            0.013$ & $0.007 \pm 0.007$ & $0.002 \pm 0.000$ & $0.002 \pm 0.000$\\
        Combined KRR & $0.006 \pm 0.008$ & $0.009 \pm 0.010$ & $0.002 \pm
            0.002$ & $0.001 \pm 0.000$ & $0.001 \pm 0.000$ & $0.001 \pm 0.000$\\
        CDM & $0.004 \pm 0.002$ & $0.007 \pm 0.001$ & $0.004 \pm 0.002$ &
            $0.001 \pm 0.000$ & $0.001 \pm 0.000$ & $0.012 \pm 0.002$\\
        Offset KS & $0.003 \pm 0.001$ & $0.002 \pm 0.001$ & $0.002 \pm
            0.000$ & $0.002 \pm 0.000$ & $0.002 \pm 0.000$ & $0.001 \pm 0.000$\\
        Offset KRR & $0.002 \pm 0.001$ & $\textbf{0.001}~\mathbf{\pm}~\textbf{0.000}$
            & $\textbf{0.000}~\mathbf{\pm}~\textbf{0.000}$
            & $\textbf{0.000}~\mathbf{\pm}~\textbf{0.000}$
            & $\textbf{0.000}~\mathbf{\pm}~\textbf{0.000}$
            & $\textbf{0.000}~\mathbf{\pm}~\textbf{0.000}$\\
        Scale KS & $0.004 \pm 0.002$ & $0.003 \pm 0.001$ & $0.002 \pm
            0.001$ & $0.002 \pm 0.000$ & $0.002 \pm 0.000$ & $0.002 \pm 0.000$\\
        Scale KRR & $\textbf{0.001}~\mathbf{\pm}~\textbf{0.000}$ &
            $\textbf{0.001}~\mathbf{\pm}~\textbf{0.000}$
            & $\textbf{0.000}~\mathbf{\pm}~\textbf{0.000}$
            & $\textbf{0.000}~\mathbf{\pm}~\textbf{0.000}$
            & $\textbf{0.000}~\mathbf{\pm}~\textbf{0.000}$
            & $\textbf{0.000}~\mathbf{\pm}~\textbf{0.000}$\\
        \bottomrule
    \end{tabular}%
    }
    \caption{1 standard deviation intervals for the mean squared errors of
        various algorithms when transferring from kin-8nh to kin-8fm.
        The values in bold are the best errors for each $n_{ta}$.}
    \label{tab:kin_8nh_to_8fm}
\end{table*}

%% file: transfer.bbl
\begin{thebibliography}{37}
\providecommand{\natexlab}[1]{#1}
\providecommand{\url}[1]{\texttt{#1}}
\expandafter\ifx\csname urlstyle\endcsname\relax
  \providecommand{\doi}[1]{doi: #1}\else
  \providecommand{\doi}{doi: \begingroup \urlstyle{rm}\Url}\fi

\bibitem[Ben-David and Urner(2013)]{ben2013domain}
Shai Ben-David and Ruth Urner.
\newblock Domain adaptation as learning with auxiliary information.
\newblock In \emph{New Directions in Transfer and Multi-Task-Workshop@ NIPS},
  2013.

\bibitem[Ben-David et~al.(2007)Ben-David, Blitzer, Crammer, and
  Pereira]{ben2007analysis}
Shai Ben-David, John Blitzer, Koby Crammer, and Fernando Pereira.
\newblock Analysis of representations for domain adaptation.
\newblock \emph{Advances in neural information processing systems},
  19:\penalty0 137, 2007.

\bibitem[Blitzer et~al.(2008)Blitzer, Crammer, Kulesza, Pereira, and
  Wortman]{blitzer2008learning}
John Blitzer, Koby Crammer, Alex Kulesza, Fernando Pereira, and Jennifer
  Wortman.
\newblock Learning bounds for domain adaptation.
\newblock In \emph{Advances in neural information processing systems}, pages
  129--136, 2008.

\bibitem[Bousquet and Elisseeff(2002)]{bousquet2002stability}
Olivier Bousquet and Andr{\'e} Elisseeff.
\newblock Stability and generalization.
\newblock \emph{Journal of Machine Learning Research}, 2\penalty0
  (Mar):\penalty0 499--526, 2002.

\bibitem[Carroll et~al.(2006)Carroll, Ruppert, Stefanski, and
  Crainiceanu]{carroll2006measurement}
Raymond~J Carroll, David Ruppert, Leonard~A Stefanski, and Ciprian~M
  Crainiceanu.
\newblock \emph{Measurement error in nonlinear models: a modern perspective}.
\newblock CRC press, 2006.

\bibitem[Cortes and Mohri(2011)]{cortes2011domain}
Corinna Cortes and Mehryar Mohri.
\newblock Domain adaptation in regression.
\newblock In \emph{Algorithmic Learning Theory}, pages 308--323. Springer,
  2011.

\bibitem[Cortes and Mohri(2014)]{cortes2014domain}
Corinna Cortes and Mehryar Mohri.
\newblock Domain adaptation and sample bias correction theory and algorithm for
  regression.
\newblock \emph{Theoretical Computer Science}, 519:\penalty0 103--126, 2014.

\bibitem[Cortes et~al.(2015)Cortes, Mohri, and
  Mu{\~n}oz~Medina]{cortes2015adaptation}
Corinna Cortes, Mehryar Mohri, and Andr{\'e}s Mu{\~n}oz~Medina.
\newblock Adaptation algorithm and theory based on generalized discrepancy.
\newblock In \emph{Proceedings of the 21th ACM SIGKDD International Conference
  on Knowledge Discovery and Data Mining}, pages 169--178. ACM, 2015.

\bibitem[Craig(1933)]{craig1933tchebychef}
Cecil~C Craig.
\newblock On the tchebychef inequality of bernstein.
\newblock \emph{The Annals of Mathematical Statistics}, 4\penalty0
  (2):\penalty0 94--102, 1933.

\bibitem[Fei-Fei et~al.(2006)Fei-Fei, Fergus, and Perona]{fei2006one}
Li~Fei-Fei, Rob Fergus, and Pietro Perona.
\newblock One-shot learning of object categories.
\newblock \emph{IEEE transactions on pattern analysis and machine
  intelligence}, 28\penalty0 (4):\penalty0 594--611, 2006.

\bibitem[Huang et~al.(2006)Huang, Gretton, Borgwardt, Sch{\"o}lkopf, and
  Smola]{huang2006correcting}
Jiayuan Huang, Arthur Gretton, Karsten~M Borgwardt, Bernhard Sch{\"o}lkopf, and
  Alex~J Smola.
\newblock Correcting sample selection bias by unlabeled data.
\newblock In \emph{Advances in neural information processing systems}, pages
  601--608, 2006.

\bibitem[Kpotufe and Garg(2013)]{kpotufe2013adaptivity}
Samory Kpotufe and Vikas Garg.
\newblock Adaptivity to local smoothness and dimension in kernel regression.
\newblock In \emph{Advances in Neural Information Processing Systems}, pages
  3075--3083, 2013.

\bibitem[Kuzborskij and Orabona(2013)]{kuzborskij2013stability}
Ilja Kuzborskij and Francesco Orabona.
\newblock Stability and hypothesis transfer learning.
\newblock In \emph{ICML (3)}, pages 942--950, 2013.

\bibitem[Kuzborskij and Orabona(2016)]{kuzborskij2016fast}
Ilja Kuzborskij and Francesco Orabona.
\newblock Fast rates by transferring from auxiliary hypotheses.
\newblock \emph{Machine Learning}, pages 1--25, 2016.

\bibitem[Kuzborskij et~al.(2013)Kuzborskij, Orabona, and
  Caputo]{kuzborskij2013n}
Ilja Kuzborskij, Francesco Orabona, and Barbara Caputo.
\newblock From n to n+ 1: Multiclass transfer incremental learning.
\newblock In \emph{Proceedings of the IEEE Conference on Computer Vision and
  Pattern Recognition}, pages 3358--3365, 2013.

\bibitem[Kuzborskij et~al.(2016)Kuzborskij, Orabona, and
  Caputo]{kuzborskij2016scalable}
Ilja Kuzborskij, Francesco Orabona, and Barbara Caputo.
\newblock Scalable greedy algorithms for transfer learning.
\newblock \emph{Computer Vision and Image Understanding}, 2016.

\bibitem[Liu et~al.(2016)Liu, Tao, Song, and Maybank]{liu2016algorithm}
Tongliang Liu, Dacheng Tao, Mingli Song, and Stephen Maybank.
\newblock Algorithm-dependent generalization bounds for multi-task learning.
\newblock \emph{IEEE transactions on pattern analysis and machine
  intelligence}, 2016.

\bibitem[Mansour et~al.(2009)Mansour, Mohri, and
  Rostamizadeh]{mansour2009domain}
Yishay Mansour, Mehryar Mohri, and Afshin Rostamizadeh.
\newblock Domain adaptation: Learning bounds and algorithms.
\newblock \emph{arXiv preprint arXiv:0902.3430}, 2009.

\bibitem[Mohri and Medina(2012)]{mohri2012new}
Mehryar Mohri and Andres~Munoz Medina.
\newblock New analysis and algorithm for learning with drifting distributions.
\newblock In \emph{Algorithmic Learning Theory}, pages 124--138. Springer,
  2012.

\bibitem[Nuske et~al.(2014)Nuske, Gupta, Narasimhan, and
  Singh]{nuske2014modeling}
Stephen Nuske, Kamal Gupta, Srinivasa Narasimhan, and Sanjiv Singh.
\newblock Modeling and calibrating visual yield estimates in vineyards.
\newblock In \emph{Field and Service Robotics}, pages 343--356. Springer, 2014.

\bibitem[Orabona et~al.(2009)Orabona, Castellini, Caputo, Fiorilla, and
  Sandini]{orabona2009model}
Francesco Orabona, Claudio Castellini, Barbara Caputo, Angelo~Emanuele
  Fiorilla, and Giulio Sandini.
\newblock Model adaptation with least-squares svm for adaptive hand
  prosthetics.
\newblock In \emph{Robotics and Automation, 2009. ICRA'09. IEEE International
  Conference on}, pages 2897--2903. IEEE, 2009.

\bibitem[Rasmussen et~al.(1996)Rasmussen, Neal, Hinton, van Camp, Revow,
  Ghahramani, Kustra, and Tibshirani]{rasmussen1996delve}
Carl~Edward Rasmussen, Radford~M Neal, Georey Hinton, Drew van Camp, Michael
  Revow, Zoubin Ghahramani, Rafal Kustra, and Rob Tibshirani.
\newblock Delve data for evaluating learning in valid experiments.
\newblock \emph{URL http://www. cs. toronto. edu/~ delve}, 1996.

\bibitem[Steinwart et~al.(2009)Steinwart, Hush, and
  Scovel]{steinwart2009optimal}
Ingo Steinwart, Don~R Hush, and Clint Scovel.
\newblock Optimal rates for regularized least squares regression.
\newblock In \emph{COLT}, 2009.

\bibitem[Stroop(1935)]{stroop1935studies}
J~Ridley Stroop.
\newblock Studies of interference in serial verbal reactions.
\newblock \emph{Journal of experimental psychology}, 18\penalty0 (6):\penalty0
  643, 1935.

\bibitem[Sugiyama et~al.(2008)Sugiyama, Nakajima, Kashima, Buenau, and
  Kawanabe]{sugiyama2008direct}
Masashi Sugiyama, Shinichi Nakajima, Hisashi Kashima, Paul~V Buenau, and
  Motoaki Kawanabe.
\newblock Direct importance estimation with model selection and its application
  to covariate shift adaptation.
\newblock In \emph{Advances in neural information processing systems}, pages
  1433--1440, 2008.

\bibitem[Tommasi et~al.(2010)Tommasi, Orabona, and Caputo]{tommasi2010safety}
Tatiana Tommasi, Francesco Orabona, and Barbara Caputo.
\newblock Safety in numbers: Learning categories from few examples with multi
  model knowledge transfer.
\newblock In \emph{Computer Vision and Pattern Recognition (CVPR), 2010 IEEE
  Conference on}, pages 3081--3088. IEEE, 2010.

\bibitem[Verstynen(2014)]{verstynen2014organization}
Timothy~D Verstynen.
\newblock The organization and dynamics of corticostriatal pathways link the
  medial orbitofrontal cortex to future behavioral responses.
\newblock \emph{Journal of neurophysiology}, 112\penalty0 (10):\penalty0
  2457--2469, 2014.

\bibitem[Vovk(2013)]{vovk2013kernel}
Vladimir Vovk.
\newblock Kernel ridge regression.
\newblock In \emph{Empirical Inference}, pages 105--116. Springer, 2013.

\bibitem[Wang and Schneider(2014)]{wang2014flexible}
Xuezhi Wang and Jeff Schneider.
\newblock Flexible transfer learning under support and model shift.
\newblock In \emph{Advances in Neural Information Processing Systems}, pages
  1898--1906, 2014.

\bibitem[Wang and Schneider(2015)]{wang2015generalization}
Xuezhi Wang and Jeff Schneider.
\newblock Generalization bounds for transfer learning under model shift.
\newblock 2015.

\bibitem[Wang et~al.(2016)Wang, Oliva, Schneider, and
  P{\'o}czos]{wang2016nonparametric}
Xuezhi Wang, Junier~B Oliva, Jeff Schneider, and Barnab{\'a}s P{\'o}czos.
\newblock Nonparametric risk and stability analysis for multi-task learning
  problems.
\newblock In \emph{25th International Joint Conference on Artificial
  Intelligence (IJCAI)}, volume~1, page~2, 2016.

\bibitem[Wasserman(2006)]{wasserman2006all}
Larry Wasserman.
\newblock \emph{All of nonparametric statistics}.
\newblock Springer Science \& Business Media, 2006.

\bibitem[Yang et~al.(2007)Yang, Yan, and Hauptmann]{yang2007cross}
Jun Yang, Rong Yan, and Alexander~G Hauptmann.
\newblock Cross-domain video concept detection using adaptive svms.
\newblock In \emph{Proceedings of the 15th ACM international conference on
  Multimedia}, pages 188--197. ACM, 2007.

\bibitem[Yu and Szepesv{\'a}ri(2012)]{yu2012analysis}
Yaoliang Yu and Csaba Szepesv{\'a}ri.
\newblock Analysis of kernel mean matching under covariate shift.
\newblock \emph{arXiv preprint arXiv:1206.4650}, 2012.

\bibitem[Zhang et~al.(2013)Zhang, Muandet, and Wang]{zhang2013domain}
Kun Zhang, Krikamol Muandet, and Zhikun Wang.
\newblock Domain adaptation under target and conditional shift.
\newblock In \emph{Proceedings of the 30th International Conference on Machine
  Learning (ICML-13)}, pages 819--827, 2013.

\bibitem[Zhang(2015)]{zhang2015multi}
Yu~Zhang.
\newblock Multi-task learning and algorithmic stability.
\newblock In \emph{AAAI}, volume~2, pages 6--2, 2015.

\bibitem[Zhou(2008)]{zhou2008derivative}
Ding-Xuan Zhou.
\newblock Derivative reproducing properties for kernel methods in learning
  theory.
\newblock \emph{Journal of computational and Applied Mathematics}, 220\penalty0
  (1):\penalty0 456--463, 2008.

\end{thebibliography}
